\documentclass{article}

\usepackage{arxiv}

\usepackage[utf8]{inputenc} 
\usepackage[T1]{fontenc}    
\usepackage{hyperref}       
\usepackage{url}            
\usepackage{booktabs}       
\usepackage{amsfonts}       
\usepackage{nicefrac}       
\usepackage{microtype}      
\usepackage{lipsum}

\usepackage{amsmath,dsfont,mathrsfs,color,verbatim}
\usepackage[numbers]{natbib}
\usepackage{multicol}
\usepackage{multirow}
\usepackage{float}
\usepackage{soul}
\usepackage{cancel}
\usepackage{graphicx,amssymb,subfig,icomma}
\newcommand{\norm}[1]{\left\lVert#1\right\rVert}

\title{Determinantal consensus clustering}

\author{
  Serge Vicente \\
  Department of Mathematics and Statistics\\
  Université de Montréal\\
  Montréal, Québec, Canada \\
  \texttt{s.vicente@umontreal.ca} \\
   \And
 Alejandro Murua \\
  Department of Mathematics and Statistics\\
  Université de Montréal\\
  Montréal, Québec, Canada \\
  \texttt{alejandro.murua@umontreal.ca} \\
}

\begin{document}
\maketitle

\begin{abstract}
Random restart of a given algorithm produces many partitions to yield a consensus clustering.
    Ensemble methods such as consensus clustering have been recognized as more robust approaches
    for data clustering than single clustering algorithms.
    We propose the use of determinantal point processes or DPP for the random restart of clustering algorithms
    based on initial sets of center points, such as $k$-medoids or $k$-means.
    The relation between DPP and kernel-based methods makes DPPs suitable to describe and quantify similarity between objects.
    DPPs favor diversity of the center points within subsets. So, subsets with more similar points have less chances of being generated
    than subsets with very distinct points.
    The current and most popular sampling technique is sampling center points uniformly at random. We show through extensive simulations
    that, contrary to DPP, this technique fails both to ensure diversity, and to obtain a good coverage of all data facets.
    These two properties of DPP are key to make DPPs achieve good performance with small ensembles.
    Simulations with artificial datasets and applications to real datasets show that determinantal consensus clustering outperform classical algorithms such as $k$-medoids and $k$-means consensus clusterings which are based on uniform random sampling of center points.
\end{abstract}


\section{Introduction}
A classical core procedure in fields such as biology, psychology, medicine, marketing, computer vision and remote sensing is to group elements based on similar features (cluster analysis) to provide a framework for learning, as reported by \cite{Jain1988}. Some clustering techniques, such as the standard $k$-means algorithm or the partitioning around medoids algorithm, are characterized by an initial choice of a subset of random points. We find the same type of initial choice in some classification techniques, such as neural networks or machine learning.
Selecting a subset of points simply at random does not take into account the diversity among the selected points, because this sort of sampling mechanism gives to every point
an equal probability of being selected.
Many similar points may be chosen simultaneously, conveying much redundancy and little representability
of the data.
In some domains of research, the diversity of the selected points is a major concern.
Ensuring diversity, so as to obtain a good coverage of all data facets, would certainly fail
when doing simple random sampling.
In contrast, determinantal point processes, or DPPs for short,
induce negative correlations between similar points \citep{Borodin2000}.
\cite{Kulesza2012} emphasize that the strength of those negative correlations
is to assign higher probability to sets of points that are more diverse.
Consequently, similar points have less chance of appearing together.
This property has established DPPs use in machine learning as models for subset selection \citep{Hafiz2013}.

The origins of DPP date back to \cite{Macchi1975} in quantum physics, where it is known as  the fermion process.
The name \emph{Determinantal Point Process} was established in Mathematics by \cite{Borodin2000}.
It also arises in studies of non-intersecting random paths \citep{Daley2003}, random spanning trees \citep{Borodin2003}, and eigenvalues of random matrices \citep{BenHough2006}. Currently, \cite{Kulesza2012} and \cite{Lavancier2015} represent central references concerning DPP.

 Traditionally, algorithms and methods of cluster analysis are gathered in two families of techniques:
hierarchical techniques and partitioning or non-hierarchical techniques.
Hierarchical techniques include agglomerative clustering with single linkage, \citep{Florek1951}, and Hidden Markov Models agglomerative clustering, \citep{Smyth1997}.
 Partitioning techniques include the partitioning around medoids (PAM) algorithm, \citep{Kaufmann1987}, and the $k$-means algorithm, \citep{Lloyd19822}.
 Recently, and due to the contribution of advanced computational methods, additional families of clustering techniques may be considered \citep{Han2011}. Among others, one finds
 probabilistic model-based techniques, which assume that each observed cluster represents a sample drawn from a specific probability distribution and, consequently, that the overall distribution of the data consists in a mixture of several distributions
 \citep{Banfield1993,Fraley1998}; density-based techniques, which model clusters as dense regions of objects in space (with respect to a local density measure) separated by sparse or low-density regions \citep{Ester1996,Ankerst1999,Stuetzle2010,Stuetzle2003};
 and grid-based techniques, which split the space into a finite number of cells that establishes a grid structure, where all the clustering process is performed \citep{Wang1997,Hinneburg1999}.

 The majority of the clustering methods seeks to obtain a single and individual optimal partition of the data, according to some internal clustering criterion, based on the principle of maximizing both within-cluster similarity and between-cluster dissimilarity.
 However, as stressed by \cite{Vega2011}, if different clustering techniques are applied to the same data, they can produce very different clustering results, due in part to a lack of an external objective and impartial criterion. The techniques' dependency on the initial choice of points can also explain those differences.
 In order to improve the quality and robustness of clustering results, \cite{Blatt1996} and \cite{Blatt1997} introduced
 a cluster-membership probabilistic framework for clustering based on physical properties of ferromagnetic models.
 Later, \cite{Strehl2002} formalized this approach, defining the \emph{cluster ensembles} framework whose main objective is to combine different clustering results into a single consolidated clustering.
 \cite{Vega2011} established four desirable properties that should be present in the results of any cluster ensemble method.
 These are (i) robustness, so that the single consolidated clustering must have better average performance than single and individual clustering algorithms; (ii) consistency, in the sense that the single consolidated clustering should produce similar results to those of all combined individual clustering algorithms;
(iii) novelty, that is, any cluster ensemble method should produce clustering solutions usually not attainable by single clustering algorithms; and  (iv) stability, in the sense that the results of the single consolidated clustering should have lower sensitivity to noise, outliers and initial conditions.
 One of the most well-known cluster ensembles methods was introduced by \cite{Monti2003}, in genomic studies and gene expression data, inspired by resampling and cross-validation techniques such as bootstrapping.

  Consensus clustering is defined as a method meant for attaining a single consolidated clustering from multiple runs of the same clustering algorithm. The obtained single consolidated clustering, built over some agreement among the several runs, represents a partition of data.
 Although not required,
 multiple runs of the algorithm could be initialized with a random restart.

 In this paper, we focus on partitioning techniques with random initial conditions.
 We explore the determinantal point process presented by \cite{BenHough2006} and \cite{Kulesza2012} for sampling the initial cluster centers.
 The use of the determinantal point process implies the choice of a real,
 symmetric and positive semidefinite matrix that measures similarity between all elements.
 The properties of this type of matrices open a connection with the well-known kernel-based methods,
 which have been widely used in pattern analysis, classification and clustering \citep{Howley2006}.
 One of the most popular kernels, and the one we use in this paper, is the Radial Basis Function, also known as the Gaussian kernel.

 The paper is organized as follows: in Section~\ref{sec:consensus1}, we introduce some basic notation, and
 describe the basic ideas related to consensus clustering.
In Section~\ref{sec:DPP}, we present the basic properties of
determinantal point processes, and put them into context as a sampling method to generate cluster centers in a partitioning clustering algorithm.
Since the choice of a proper kernel is central in the construction of determinantal point processes,
kernel-based methods are also introduced in this section. In Section~\ref{sec:consensus:DPP},
we introduce our proposed methodology for determinantal consensus clustering, or \emph{consensus DPP,} for short.
In Section~\ref{sec:performance}, we perform an extensive simulation in order to evaluate the performance of consensus DPP.
The performance of the proposed algorithm on real datasets is presented in Section~\ref{sec:real1}.
A comparison with other partitioning methods is also shown in these last two sections.
We conclude with a few thoughts and a discussion in Section~\ref{sec:conclusions1}.

\section{Consensus clustering}\label{sec:consensus1}

Throughout the paper the data will be
denoted by $\mathcal{S}=\{{x_1}, \dots, {x_n}\}\subset \mathds{R}^p$, where $x_i$ represents a $p$-dimensional vector, for $i=1,\dots,n$ and $n \geq 2$.
Cluster analysis consists in a range of algorithms and methods that divide a discrete set of elements into several subsets, or clusters, sharing some common features or properties.
The process of division into subsets follows two criteria:
(a) Division is exclusive, i.e., subsets do not overlap, forming a partition $\{C_1, C_2, \dots, C_K\}$ of $\mathcal{S}$.
That is $\mathcal{S} = \bigcup_{i=1}^K C_i$, and $C_i \cap C_j = \emptyset$ whenever $i\not= j$, $i,j = 1, \dots, K.$
(b) Division is intrinsic or unsupervised, i.e., the division is based only on a proximity matrix, rather than using category labels denoting an a priori partition.

 Consider
 a particular partitioning clustering technique run $R$ times on the data $\mathcal{S}$. The agreement among the several runs of the algorithm is based on the \emph{consensus matrix} $C$. This is a $n\times n$ symmetric matrix whose entries $\{ C_{ij}, \; i,j=1,\dots,n\}$ represent the proportion of runs in which elements $x_i$ and $x_j$ of $\mathcal{S}$ fall in the same cluster.
Let $r$ represent a specific run of the clustering algorithm, $r=1,\dots,R,$ and let $C_r$ be the associated $n\times n$ symmetric binary matrix with entries
\begin{equation}\label{binary}
  c^{r}_{ij}=\left\{
  \begin{array}{ll}
    1, & \hbox{if $x_i$ and $x_j$ belong to the same cluster;} \\
    0, & \hbox{otherwise,}
  \end{array}
\right.
\end{equation}
for $i,j=1,\dots,n$. The consensus matrix $C$ is then the $n\times n$ symmetric matrix with entries defined by
\begin{equation}\label{equation2}
  C_{ij}=\sum_{r=1}^{R}c^{r}_{ij} / R,
\end{equation}
for $i,j=1,\dots,n$. The entry $C_{ij}$ is known as \emph{consensus index}. Obviously, the diagonal entries are given by $C_{ii}=1,$ for $i=1,\dots,n$.\\
Clusters can also be defined using graph theory by considering the graph $(V,E)$ whose vertices are given by the elements $x_i\in\mathcal{S}$. The set of edges $E$ is defined by connecting each pair of elements sharing some common features or properties.
A clustering configuration with $K$ clusters consists in an undirected graph with $K$ connected components.
\cite{Murua2008} established that the single consolidated clustering with $K$ final clusters obtained by consensus clustering represent the $K$ connected components of the \emph{consensus graph}. That is the graph over the observations with an edge between any pair of elements that belong to the same cluster in the majority of the configurations.
The majority concept is tied to the consensus index defined in \eqref{equation2}.

\section{The determinantal point process}\label{sec:DPP}

Keeping the notation from the previous section,
a DPP is a probability measure on $2^{\mathcal{S}}$ that assigns probability
\begin{equation}\label{definition22}
  P\left(Y\right)= \det(L_Y) / \det(L+I_n),
\end{equation}
to any subset $Y\in 2^{\mathcal{S}}$,
where
$L$ is a $n\times n$ real, symmetric and positive semidefinite matrix that measures similarity between all pairs of elements of $\mathcal{S}$; $L_Y$ is the principal submatrix of $L$ whose rows and columns are indexed by $Y$, i.e., $L_Y=\left(L_{ij}\right)_{i,j\in Y}$; and $I_n$ is the $n\times n$ identity matrix. If $\boldsymbol{Y}$ is the random variable that represents the subset selected from $2^{\mathcal{S}}$,
then we write $\boldsymbol{Y}\sim DPP_{\mathcal{S}}(L)$ for the corresponding determinantal process.
The matrix $L$ is known as the kernel matrix of the DPP \citep{Kulesza2012,Kang20132,Hafiz2014}.
It can be shown \citep{Kulesza2012} that $\det(L+I_n) = \sum_{Y \subset 2^\mathcal{S}} \det(L_Y)$, hence
\eqref{definition22} does indeed define a probability mass function over all subsets in $2^\mathcal{S}$.
This definition states restrictions on all the principal minors of the kernel matrix $L$, denoted by $\det(L_Y)$.
Indeed, as $P\left(\boldsymbol{Y}=Y\right)\propto\det(L_Y)$ represents a probability measure, we have $\det(L_Y)\geq0$, for any $Y\subseteq \mathcal{S}$.
This implies that any symmetric positive semidefinite matrix can be taken as kernel matrix $L$, where its eigenvalues are such that $\lambda_i(L)\geq 0$, for every $i=1, \ldots,n$.

\subsection{Relation between kernel-based methods and DPP}\label{sec:mercer}

Determinants have a well-known geometric interpretation. Because $L$ is positive semidefinite,
it can be decomposed as $L=B^TB$, where $B$ is a $m\times n$ matrix.
Denoting the columns of $B$ by $B_i$, for $i =1,\dots,n$, we have
\begin{equation}\label{volume}
  P\left(\boldsymbol{Y}=Y\right)\propto\det\left(L_Y\right)=\mbox{Vol}^2\left(\left\{B_i\right\}_{i\in Y}\right),
\end{equation}
where Vol$^2$ represents the squared volume of the parallelepiped spanned by the columns of $B$ corresponding to elements in $Y$. The columns of $B$ can be interpreted as feature vectors describing the elements of $\mathcal{S}$ and, therefore, $L$ measures similarity using dot products between feature vectors. As the dot product is the most natural similarity measure between vectors, it establishes a connection with the well-known kernel-based methods. Kernels are widely used
to describe and quantify how any two objects are related.
They have been widely used in pattern analysis, like classification and clustering \citep{Howley2006}.
By \eqref{volume}, we can see that the probability assigned by a DPP to a subset $Y$ is related to the volume spanned by its associated feature vectors: sets composed of very diverse elements have higher probabilities, because their feature vectors are more orthogonal, and span larger volumes.

Considering clustering analysis, \cite{Jain1988} established that, for datasets with ellipsoidal clustered structures, ``sum-of-squares'' based methods have proved to be effective. However, if the frontiers that separate clusters are non-quadratic, these methods will fail to generate an effective clustering configuration. One of the several approaches to deal with this problem consists in nonlinearly transforming the data into a high-dimensional feature space, so that clustering analysis can be conducted in this feature space, constructing an optimal separating hyperplane \citep{Vapnik19952,Girolami2002,Murua2008}.
Let $\mathcal{H}$ be an embedding Hilbert space, and consider a mapping $\boldsymbol{\Phi}:\mathcal{S}\rightarrow\mathcal{H}$. The set containing all the transformed elements of $\mathcal{S}$ is represented by
$$\boldsymbol{\Phi}(\mathcal{S}):=\left\{\boldsymbol{\Phi}({x_1}), \dots, \boldsymbol{\Phi}({x_n})\right\}.$$
Each $x_i$ is mapped into a high-dimensional Hilbert space  $\mathcal{H}$ with coordinates $\boldsymbol{\Phi}(x_i)=\left(\phi_1(x_i),\phi_2(x_i),\dots\right),$
for $i=1,\dots,n$.
Because the feature space $\mathcal{H}$ may be of high and possibly infinite dimension, working directly with the transformed data is an unrealistic option.
Kernel-based methods calculate a similarity measure between each pair of elements on the feature space $\mathcal{H}$,
to afterwards use algorithms that only need the value of this measure \citep{Schlkopf2004}.
Since the feature space is a Hilbert space,
the inner product $\langle\boldsymbol{\Phi}(x_i),\boldsymbol{\Phi}(x_j)\rangle_{\mathcal{H}}$ is the obvious and simplest similarity measure to conceive. The algorithms of kernel-based methods are said to employ kernel functions, since the pairwise inner products $\langle\boldsymbol{\Phi}(x_i),\boldsymbol{\Phi}(x_j)\rangle_{\mathcal{H}}$ can be computed directly from the original data using an appropriate kernel function $\kappa: \mathcal{S}\times\mathcal{S} \rightarrow \mathds{R}$. That is,
  \begin{equation}\label{kfunction}
    \kappa(x_i,x_j)=\langle\boldsymbol{\Phi}(x_i),\boldsymbol{\Phi}(x_j)\rangle_{\mathcal{H}},
  \end{equation}
  for $i,j = 1,\dots, n$.
  The kernel function is then able to represent the inner products in $\mathcal{H}$ in the original space $\mathcal{S}$. Consequently, kernel-based methods replace the inner products with the kernel function, a fact known as the \emph{kernel trick}. However, \eqref{kfunction} raises the issue of which type of kernel functions are allowed.

  Mercer's Theorem \citep{Vapnik19952} tells us whether or not a function $\kappa$ is actually an inner product $\langle\boldsymbol{\Phi}(x_i),\boldsymbol{\Phi}(x_j)\rangle$ in some space ${\mathcal{H}}$.
Assume that all possible data $\mathcal{S}$ live in a compact subset $\mathcal{X}$ of $\mathds{R}^p$.
Suppose $\kappa: \mathcal{S}\times\mathcal{S} \rightarrow \mathds{R}$ is a continuous symmetric function satisfying
$$\sum_{i=1}^{n}\sum_{j=1}^{n}a_ia_j\kappa(x_i,x_j)\geq 0,$$
for any finite set of points $\left\{x_i\right\}_{i=1}^{n}$ in $\mathcal{X}$ and real numbers $\left\{a_i\right\}_{i=1}^{n}$. Furthermore, suppose that
$$\int_{\mathcal{X}\times\mathcal{X}}\kappa(x_i, x_j)f(x_i)f(x_j)dx_idx_j\geq 0,$$
for all squared-integrable functions $f(\cdot)$ on $\mathcal{X}$.
Then, $\kappa$ can be expanded in a uniformly convergent series in terms of a unique enumerable set of  non-negative eigenvalues $\left\{\lambda_r\right\}_{r=1}^{+\infty}$, and associated squared-integrable orthogonal eigenfunctions $\left\{\psi_r\right\}_{r=1}^{+\infty}$,
$$\kappa(x_i, x_j)=\sum_{r=1}^{+\infty}\lambda_r\psi_r(x_i)\psi_r(x_j).$$
Recalling \eqref{kfunction}, if a function $\kappa$ satisfies Mercer's Theorem, we can define a feature map $\boldsymbol{\Phi}:\mathcal{S}\rightarrow\mathcal{H}$ as follows:
$$\boldsymbol{\Phi}(x_i)=\left(\sqrt{\lambda_1}\psi_1(x_i),\sqrt{\lambda_2}\psi_2(x_i),\dots\right),$$
for $i=1,\dots,n$. In this case, we say that the kernel function $\kappa$ is a \emph{Mercer kernel}.
A particular property of Mercer kernels is that they also are positive semidefinite kernels.
This ensures that the $n\times n$ matrix defined by
$$\boldsymbol{K}= \left( \kappa(x_i,x_j)\right)_{i,j=1}^n = \left(\langle\boldsymbol{\Phi}(x_i),\boldsymbol{\Phi}(x_j)\rangle_{\mathcal{H}}\right)_{i,j=1}^n,$$
is positive semidefinite.
$\boldsymbol{K}$ is also a Gram matrix \citep{Horn20122}.

Kernel functions are often considered measures of similarity, since a higher kernel value represents a higher correlation in the associated Hilbert space. Therefore, for the purpose of cluster analysis and the choice of the similarity matrix $L$ for the DPP established in \eqref{definition22}, Mercer kernels are perfect candidates, so that
\begin{equation}\label{Lmatrix}
  L=\left[\kappa(x_i,x_j)\right]_{i,j=1}^n.
\end{equation}

\subsection{Choice of kernel: the radial basis function kernel}

The choice of the most appropriate kernel function is a critical step in the application of any kernel-based method. However, as pointed by \cite{Howley2006}, there is no rule or consensus about the choice of the most suitable kernel function for a particular problem. Ideally, the suitable kernel function is chosen according to prior knowledge of the problem domain
\citep{Howley2006,Lanckriet2004}, which is rarely observable in practice.
In the absence of expert knowledge, a common choice is the \emph{Radial Basis Function} (RBF) kernel (or \emph{Gaussian kernel}):
\begin{equation}\label{RBF}
  \kappa(x_i,x_j)=\exp\left(-\|x_i-x_j\|^2 / (2\sigma^2) \right),
\end{equation}
where the scale parameter $\sigma$, known as the bandwidth of the kernel, represents the relative spread of the distances $\|x_i-x_j\|$. Here, the distance $\|x_i-x_j\|$ represents the Euclidean distance between $x_i$ and $x_j$, a common choice for the RBF, which we will also follow. The RBF is a Mercer kernel, as presented in Section~\ref{sec:mercer}, and details concerning its expansion in terms of non-negative eigenvalues $\left\{\lambda_r\right\}_{r=1}^{+\infty}$ and associated eigenfunctions $\left\{\psi_r\right\}_{r=1}^{+\infty}$ can be found in \cite{Fausshauer2011}.
 This particular kernel has been extensively used in many studies, due to its appealing mathematical properties, as mentioned by \cite{Girolami2002}. A particular property of the Gaussian kernel is that it is positive and bounded from above by one,
 making it directly interpretable as a scaled measure of similarity between $x_i$ and $x_j$.

 For the purpose of this paper, we choose the Gaussian kernel defined in \eqref{RBF} for building the similarity matrix $L$ of the DPP with \eqref{Lmatrix}. The computation of the RBF kernel requires the estimation of the bandwidth parameter $\sigma$. As pointed by \cite{Murua2014}, most of the literature considers $\sigma$ as a parameter that can be estimated by observed data.  Inspired by \cite{Blatt1996} and \cite{Blatt1997}, we estimate $\sigma^2$ by the average of all pairwise squared Euclidean distances, i.e.,
\begin{equation}\label{sigmahat}
  \widehat{\sigma}^2= 2 \sum_{i<j}\|(x_i-x_j)\|^2 / \bigl( n(n-1)\bigr).
\end{equation}
\noindent The authors justify the use of the average to estimate $\sigma^2$ based on local structure of the data and identification of high-density regions in the data space. Other methods of estimating the bandwidth parameter can be found in the literature. \cite{Murua2008} do not consider $\sigma$ as fixed and propose an adaptive bandwidth selection procedure, where $\sigma$ depends on the data points. They explore the relationship between Potts model and kernel density estimation, building an algorithm based on Markov Chain Monte Carlo methods to obtain a Bayesian estimate of $\sigma$. \cite{Murua2014} consider $\sigma$ as fixed and obtain its Bayesian estimate based on the Wang-Landau algorithm \citep{wang2001efficient}. \cite{sejdinovic2013equivalence} refer the median of the pairwise Euclidean distances as a common choice. However, \cite{chaudhuri2017mean} demonstrate that the use of the average and the median of Euclidean distances to estimate $\sigma$ produce similar clustering results for the majority of situations. They justify the use of the average distances by its simplicity and fast computation even when the dataset is large. We decided to use the average for the same reasons.

\noindent To explore the sensitivity of the clustering configuration to the parameter $\sigma$, we decided to introduce a tuning parameter $s>0$, which will be estimated heuristically by simulation in Section~\ref{sec:simulation}. With the bandwidth estimate given by \eqref{sigmahat} and the tuning parameter $s$, the RBF kernel in \eqref{RBF} will be adjusted to
\begin{equation}\label{gausskern}
  \kappa(x_i,x_j)=\exp\left(-\|(x_i-x_j)\|^2 / ( 2s\widehat{\sigma}^2)\right).
\end{equation}

\section{Consensus DPP}
\label{sec:consensus:DPP}

In this section we develop a partitioning clustering algorithm that will be run $R$ times over the set $\mathcal{S}$, in order to obtain a consolidated clustering configuration by consensus clustering.
To build a consensus clustering, any partitioning clustering method can be chosen.
We propose the use of determinantal point processes as the partition generating algorithm.
The algorithm is also based on a Voronoi diagram as described next.

Voronoi diagrams support many clustering techniques \citep{Aurenhammer1991}, such as the $k-$means and $k$-medoids algorithms, for example. A Voronoi diagram refers to a partition of the space into several cells or regions, based on a subset of elements that are called \emph{generator points}, or simply \emph{generators}. Each cell includes only one generator and all the space points that are closer to that generator than to any other generator.  For a formal definition of Voronoi diagram, see for example \citep{Okabe2000}.

Let $\mathcal{P}=\{{p_1}, \dots, {p_K} \}\subseteq \mathcal{S}$ be
the generator set of the Voronoi diagram. The
$p$-dimensional Voronoi polyhedron associated with ${p_i}$,
$i= 1,\dots,K$
is the
region defined by
\begin{equation*}
  V({p_i})=\left\{{x}\in \mathcal{S} : \norm{{x}-{p_i}} < \norm{{x}-{p_j}},\, \text{\ for all \ } j\neq i, j= 1,\dots,K\right\}.
\end{equation*}
The set $\mathcal{V}(\mathcal{P})=\left\{V({p_1}),\dots, V({p_K})\right\}$
is said to be the $p$-dimensional Voronoi diagram generated by $\mathcal{P}$. We call ${p_i}$ the generator point or generator of the $i$th Voronoi polyhedron.
The Voronoi diagram is a partition of the data $\mathcal{S}$, and hence a clustering of the data.
In order to obtain a Voronoi diagram, one needs to select the set of generators.
We proposed using a determinantal point process (DPP) rather than a classical random sampling for this step.
A DPP intents to capture and model negative correlations between the elements of $\mathcal{S}$ \citep{Hafiz2013},
so that the inclusion of one element makes the inclusion of other similar elements less likely.
We conjecture that sampling from a DPP for Voronoi generators is more efficient than sampling
generator points uniformly at random as it is usually done in PAM.
Our experiments in Section~\ref{sec:simulation} corroborate this belief.

\cite{BenHough2006}
and \cite{Kulesza2012} present an efficient scheme to sample from a DPP.
The algorithm is based on the following observations.
Let $L=\sum_{i=1}^{n} \lambda_i(L){v}_i{v}_i^T$ be an orthonormal eigendecomposition of $L$.
For any set of indexes $J \subseteq \{1,2,\ldots,n\}$, define the subset of eigenvectors $V_J = \{ v_i : i \in J\}$,
and the associated matrix $\mathcal{K}_J = \sum_{i\in J} {v}_i {v}_i^T.$
It can be shown that the matrix $\mathcal{K}_J$ defines a so-called \emph{elementary} DPP which we denote by DPP$(\mathcal{K}_J)$.
It turns out that the DPP$_\mathcal{S}(L)$ is a mixture of all elementary DPP given by the index sets $J$.
That is
\[
\operatorname{DPP}_\mathcal{S}(L) = \biggl[ \sum_{J} \operatorname{DPP}(\mathcal{K}_J) \prod_{i\in J} \lambda_i(L) \biggr] / \det{(L + I_n)}
\]
The mixture weight of $\operatorname{DPP}(\mathcal{K}_J)$ is given by the product of the eigenvalues $\lambda_i(L)$ corresponding to the eigenvectors $\boldsymbol{v}_j\in V_J$, normalized by $\det\left(L + I_n\right) =\prod_{i=1}^{n}\left[\lambda_i(L)+1\right]$.
Sampling can be realized by first selecting an elementary DPP, $\operatorname{DPP}(\mathcal{K}_J)$, with probability equal to its mixture component weight, and then, in a second step,
sampling $\boldsymbol{Y}\sim \operatorname{DPP}(\mathcal{K}_J)$.
In particular, it can be shown that in this case, necessarily $\operatorname{card}(\boldsymbol{Y})=\operatorname{rank}\left(\mathcal{K}_J\right)$.

For the purpose of consensus clustering and the construction of the consensus matrix with entries defined by \eqref{equation2}, we will consider $R$ runs of the sampling algorithm of \cite{BenHough2006} and \cite{Kulesza2012}.
The sampling of the $R$ sets is done from the DPP with associated kernel matrix $L$ constructed with the RBF kernel in \eqref{gausskern}.
This yields $R$ generator sets $\{ \mathcal{P}_r\}_{r=1}^R$. For each generator set, we construct a $p$-dimensional Voronoi diagram $\mathcal{V}(\mathcal{P}_r)$ based on the similarities given by $L$.
The binary matrix $C_r$ with entries given by \eqref{binary} has an entry $c_{ij}^r = 1$ if and only if
the points $x_i$ and $x_j$ fall in the same Voronoi cell, $r=1,\dots, R$.
The consensus matrix $C$, constructed with the consensus indexes defined by \eqref{equation2}, is finally
given by the average of all the matrices $C_r$ over the $R$ runs, $r=1,\ldots, R.$

The consensus matrix represents the proportion of runs in which two elements $x_i$ and $x_j$ of $\mathcal{S}$ belong to the same cluster. The consolidated clustering configuration is obtained by a thresholding procedure. According to \cite{Blatt1996}, if $C_{ij}\geq \theta$, with $0\leq\theta\leq 1$, points $x_i$ and $x_j$ are defined as ``friends'' and then included in the same final cluster. Moreover, all mutual friends (including friends of friends, etc.) are assigned to the same cluster. It can be shown that this is equivalent to finding the connected components of the consensus graph
introduced in Section~\ref{sec:consensus1} \citep{Murua2008}.

The choice of the threshold is not an easy task.
Although a value of $\theta=0.5$ makes sense most of the time, it might not be the optimal choice.
In fact, \cite{Murua2014} shows that choosing a fixed and unique threshold does not necessarily give the best clustering results.
Changing the threshold yields different clustering results. Many of those clusterings are worth exploring.
\cite{Murua2014} consider all threshold values from the set of all different observed consensus indexes $C_{ij}$ (see in \eqref{equation2}).
If there are $t$ different consensus indexes, we will have a collection of $t$ thresholds $\theta_1, \theta_2, \dots, \theta_t$.
For each threshold $\theta_i, \; i=1,\dots,t$, a consolidated clustering configuration with $K(\theta_i)$ clusters is obtained.
If $\theta_i=0$, we obtain a graph with $K(0)=1$ cluster, that is, $\mathcal{S}$.
If $\theta_i=1$, we obtain a graph with $K(1) =n$ clusters; that is, each element of $\mathcal{S}$ is an isolated point and form a singleton cluster of size one.
In general, clustering configurations with one cluster or $n$ clusters are of no interest. Therefore,
thresholds $\theta_i$ that are too low or too large are not relevant.
We adopt a mixed strategy between choosing a predetermined fixed threshold \citep{Blatt1996} and
studying a sequence of interesting thresholds \citep{Murua2014}.
We consider a sequence of $t$ predetermined thresholds $\theta_1, \theta_2, \dots, \theta_t$ that are above a certain minimum threshold $\tau$.
The value of $\tau = 0.6$ has been determined through simulations. These are reported in Section~\ref{sec:simulation}.

Moreover, we are not interested in a clustering configuration with too many small clusters.
We impose a minimal size for each cluster, accepting only clustering configurations with cluster sizes larger than that minimal value. For the establishment of the minimal size, we decided to take a classical approach, inspired by the ``square-root choice'' for the number of bins of a histogram, $\sqrt{n}$. However, we also consider
a more general case that eliminates all clusters with less than $n^a$ elements for a predetermined power $a\in(0,1)$.
The optimal value of the power $a$ depends on various considerations such as the data size, the data dimension, and the number of clusters. We have studied it through the simulations reported in Section~\ref{sec:simulation}.
In summary, we examine all the $t$ consolidated clustering configurations obtained with all the different considered thresholds $\theta_1, \theta_2, \dots, \theta_t$. If one configuration does not satisfy the minimal cluster size criterion, we merge each small cluster with its closest ``large'' cluster, according to the following procedure, inspired by single linkage:
select the component $\mathcal{V}$ that has the smallest cluster size $< n^{a}$;
find the pair of indexes $(i^*, j^*) \in \{1,\ldots,n\}$ that satisfies
  $C_{i^*  j^*} = \max\{ C_{ij} : x_i \in \mathcal{V}, x_j \not\in \mathcal{V} \}$;
  merge the component $\mathcal{V}$ to the component that includes  $\boldsymbol{x_{j^*}}$;
repeat the merging procedure until there are no more connected components with cluster size
    smaller than $n^a$.
    Other linkage merging criteria are possible, such as average linkage or minimax linkage \citep{Ao2004,Bien2011}.
    However, in our experiments these two merging linkage criteria perform similarly to single linkage merging.

    As mentioned above, the choice of the power $a$, the minimum threshold $\tau$, and the number $R$ of runs of
    our clustering algorithm, have been determined via simulation (see Section~\ref{sec:simulation}).

\subsection{Choosing an optimal clustering}

Following our mixed strategy for the thresholding procedure, we end up with a set of consolidated clustering configurations that meet our minimal cluster size criterion.
One question remains: which configuration to keep as final clustering configuration.
The answer depends on the criterion chosen to  measure the adequacy of the clustering configuration.
We use kernel-based measures that depend only on the kernel matrix $L$ in order to be computed.

Consider the RBF kernel $\kappa(x_i,x_j)$ defined by \eqref{RBF}.
Compute the mean of the transformed data $\boldsymbol{\Phi}(\mathcal{S})$,
$\bar{\boldsymbol{\Phi}} = \sum_{i=1}^n \boldsymbol{\Phi}( x_i) / n$.
The \emph{mean scattering} induced by the kernel on the data \citep{Vert2004,Fan2010}  is defined as
\begin{multline*}
  V_{\mathcal{S}}  =
\sum_{i=1}^n \frac{\lVert \boldsymbol{\Phi}( x_i)  - \bar{\boldsymbol{\Phi}} \lVert }{n}
 = \frac{1}{n}\sum_{i=1}^{n}\left\{
\kappa(x_i,x_i)-\sum_{j=1}^{n}\frac{2\, \kappa(x_i,x_j)}{n}+\sum_{j, \ell=1}^{n}\frac{\kappa(x_j,x_\ell)}{n^2} \right\}^{\frac{1}{2}}.
\end{multline*}
Let $\mathcal{V}= \{\mathcal{V}_1, \ldots, \mathcal{V}_K\}$ be a cluster configuration with $K$ clusters. Also, let $n_k$ be the size of
cluster $\mathcal{V}_k$, and consider its center in the transformed space $\bar{\boldsymbol{\Phi}}_k = \sum_{i\in J_k} \boldsymbol{\Phi}(x_i)/ n_k$, where
$J_k = \{ i : x_i \in \mathcal{V}_k\}$, $k=1,\ldots,K.$
As with the mean scattering, \cite{Fan2010} define
similarly,  the kernel induced scattering within each cluster $\mathcal{V}_k$,
$W_{\mathcal{V}_k}  = \sum_{i\in J_k} \lVert \boldsymbol{\Phi}( x_i)  - \bar{\boldsymbol{\Phi}}_k \lVert / n_k$. That is,
$$
  W_{\mathcal{V}_k}
= \frac{1}{n_k}\sum_{i\in J_k}  \left\{
\kappa(x_i,x_i)-\sum_{j\in J_k}\frac{2\, \kappa(x_i,x_j)}{n_k}+\sum_{j, \ell\in J_k} \frac{\kappa(x_j,x_\ell)}{n_k^2} \right\}^{\frac{1}{2}}.
$$
The average within cluster scattering associated with clustering configuration $\mathcal{V}$ is defined as
\begin{equation*}
  W_\mathcal{V} =\sum_{k=1}^{K} W_{\mathcal{V}_k} /(K\, V_{\mathcal{S}}).
\end{equation*}
To measure the total scattering between clusters, one considers the distance between cluster means in the transformed space,
$B(\mathcal{V}_i, \mathcal{V}_j) = \lVert \bar{\boldsymbol{\Phi}}_i - \bar{\boldsymbol{\Phi}}_j \lVert$, that is,
\begin{align*}
 B^2(\mathcal{V}_i,\mathcal{V}_j) = \bigl[ B(\mathcal{V}_i,\mathcal{V}_j) \bigr]^2 & =
\sum_{i, j \in J_i} \frac{\kappa(x_i,x_j)}{n_i^2}  -
\sum_{i\in J_i}\sum_{j\in J_j}\frac{\kappa(x_i,x_j)}{n_in_j} +\sum_{i, j\in J_j} \frac{\kappa(x_i,x_j)}{n_j^2},
\end{align*}
and $B_\mathcal{V} = \sum_{i=2}^K \sum_{j=1}^{i-1} n_i n_j B(\mathcal{V}_i,\mathcal{V}_j) / \sum_{i=2}^K\sum_{j=1}^{i-1} n_i n_j$.
A simple measure of quality of the clustering configuration may be easily derived from the similarity ratio introduced by \cite{Chen-et-al-2002},
\begin{equation} \label{eq:sr}
  SR_\mathcal{V} = 1 - \bigl[n /(n-1)\bigr] \bigl[ W_\mathcal{V}  / (W_\mathcal{V} + B_\mathcal{V} )\bigr].
\end{equation}
Because a good clustering configuration must have a small within variance and a large between variance,
the larger $SR_\mathcal{V}$, the better the clustering $\mathcal{V}$.\\
An alternative measure, \emph{the kernel-based validation index,}  developed by \cite{Fan2010} is based on
\begin{equation*} 
  \tilde{B}_\mathcal{V}=(B_{\max}/B_{\min})\sum_{i=1}^{K}\sum_{j=1}^{K} \bigl[ B^{2}(\mathcal{V}_i,\mathcal{V}_j) \bigr]^{-1},
\end{equation*}
where $B_{\max} = \max_{(i,j)} B^2(\mathcal{V}_i,\mathcal{V}_j)$, and
$B_{\min} = \min_{(i,j)} B^2(\mathcal{V}_i,\mathcal{V}_j)$.
The measure is given by
\begin{equation}\label{KRV}
  KVI_\mathcal{V}= \alpha W_\mathcal{V} + \tilde{B}_\mathcal{V},
\end{equation}
where $\alpha$ is a tuning parameter that the authors set to the value of $\tilde{B}_\mathcal{V}$ associated with
the largest clustering size $K$  among those clustering configurations being considered.
The optimal clustering configuration among the set of all retained clustering configurations is the one that minimizes \eqref{KRV}.

\subsection{Setting appropriate consensus DPP parameters}\label{sec:simulation}

In this section, we will conduct simulations to choose the tuning parameters of the proposed clustering method,
these are
(i) the tuning parameter $a$
for the minimal size of clusters $n^{a}$;
(ii) the number  $R$ of sufficient runs of the clustering algorithm;
(iii) the inferior limit $\tau$ for the threshold range to obtain the consolidated clustering configurations; and
(iv) the tuning parameter $s$ in \eqref{gausskern} for the sensitivity of the bandwidth parameter $\hat \sigma^2$.

\paragraph{Data generation.}

\; The simulated data were generated with the algorithm of \cite{Melnykov2012}.
This generates datasets from $p$-variate Gaussian mixtures with $K$ components,
$\sum_{k=1}^K \pi_k \phi_p(\cdot; \mu_k, V_k)$, where $\phi_p$ denotes the $p$-variate normal density.
The mean vectors of the components, $\{{\mu}_1, \ldots,{\mu}_K\}$,
are obtained as $k$ independent realizations from a uniform $p$-variate unit hypercube.
The covariance matrix of each component, $V_k$, is obtained as a realization from the $p$-variate standard Wishart distribution with $p + 1$ degrees of freedom. The mixing proportions $\pi_k$ are generated from a Dirichlet distribution on the standard $K-1$ simplex, so that $\sum_{k=1}^{K} \pi_k =1$.
The number of elements generated from each component is obtained from the multinomial distribution based on the mixing proportions.

The algorithm allows to control the \emph{pairwise overlap} between two components, which measures the interaction between components, and controls the clustering complexity of datasets simulated from the mixtures.
For the purposes of our study, we focus on low pairwise overlap cases, as the notion of clustering itself becomes less meaningful as the overlap degree between clusters becomes large.
According to \cite{Melnykov2012}, an average overlap of 0.4 is considered extreme while an average overlap of 0.001 is considered very low.
We fix a maximum pairwise overlap of 0.01 between any two components. The values of the pairwise overlap
were then generated uniformly at random taking into account this constraint.
Moreover, the data were obtained from mixtures with ellipsoidal covariance matrices, and unequal number of elements per component.

We studied the effect of three variables on the clustering results. These are
the number of observations per dataset $n \in \{150 \text{\ (low)}, 500\text{\ (medium)}, 1500 \text{\ (large)}\}$;
the number of variables per dataset $p$: low ($2\leq p\leq7$), medium ($8\leq p\leq12$) and large ($13\leq p\leq20$);
and the number of components or clusters per dataset $G$: low ($2\leq K\leq5$), medium ($6\leq K\leq10$) and large ($11\leq K\leq20$).
The values of $p$ and $G$ were chosen randomly in each case once the level (low, medium or large) was chosen.
The three categories generate a $3^3$ factorial design with 27 experimental conditions. To simulate data for this factorial design, we ensured that no cluster with a small number of elements was present in the simulated datasets. This was done because,
according to the merging procedure described in Section~\ref{sec:consensus:DPP}, small clusters
would be inevitably merged with a larger cluster.
This consideration results in simulated data with more or less balanced clusters. However, as it is impossible to simulate a dataset with a low number of observations ($n=150$) and a large number of clusters ($11\leq K\leq20$) that contains no cluster with a small number of elements, this case has been excluded from our analysis. Therefore, we only consider 24 experimental conditions of the $3^3$ factorial design, generating 10 datasets per condition, obtaining a total of 240 datasets. We will refer to the 24 experimental conditions as the 24 experimental \emph{scenarios,} or
scenarios, for short.

\paragraph{Measuring the quality of the clustering.}

\; In order to measure the quality of the clustering results we use the
Adjusted Rand Index (ARI), which is a common measure
of goodness-of-fit in the clustering literature \citep{Yeung-et-al-2001,Murua2014}.
The ARI was
first introduced by \cite{Rand1971} and later adjusted for randomness by \cite{Hubert1985}. It is a measure of agreement between two clustering configurations. The original Rand Index counts the proportion of elements that are either in the same clusters in both clustering configurations or in different clusters in both configurations. The adjusted version of the Rand Index corrected the calculus of the proportion, so that its expected value is zero when the clustering configurations are random. The larger the ARI, the more similar the two configurations are, with the maximum ARI score of 1.0 indicating a perfect match.

We will also use the relative difference in the estimated number of clusters and the true number of clusters (RN) presented by \cite{Murua2018}:
$$ \operatorname{RN}= (\sqrt{\widehat{G}}-\sqrt{G} ) / \sqrt{G},$$
where $\widehat{G}$ is the estimated number of clusters and $G$ is the true number of clusters. Here, small absolute values of RN are preferred and, therefore, the absolute value of RN will be used as another measure of clustering quality.

\paragraph{Results.}

\; For each of the ten replicas associated with one of the 24 scenarios, we run consensus DPP with $R=1000$,
and $a \in \{\nicefrac{1}{4},\nicefrac{1}{3},\nicefrac{1}{2},\nicefrac{2}{3}\}$.
To obtain the consolidated clustering configurations with the thresholding procedure,
we set the threshold to a unique fixed value in $\{0.4, 0.5, 0.6, 0.7, 0.8, 0.9\}$.
The quality of each clustering configuration associated with each threshold was assessed with the ARI criterion,
after the first 10, 50, 100, 200, 300, 500, 700, 900 and 1000 runs.
As each case is evaluated on the ten replica datasets of a particular scenario,
the ARIs were averaged over the ten runs. For the choice of the optimal clustering, we used the similarity ratio and the kernel-based validation index, defined in \eqref{eq:sr} and \eqref{KRV}, respectively.
After comparing the results, we decided to only keep the kernel-based validation index since it provides better results.

From the results (not shown here) we observe a relative stability or the ARI mean values for most of the cases.
Also, setting $a=\nicefrac{1}{2}$ for the minimal cluster size criterion of the merging procedure seems a reasonable choice,
which corresponds to the classic choice of $\sqrt{n}$.
With respect to the number of runs required to achieve a good clustering fit, the simulations show
that  a number of runs between 50 and 200 is adequate.
As the best results are obtained most of the times with 200 runs, we adopt $R=200$.
The choice of the inferior limit $\tau$ for the range of thresholds depends slightly on the experimental scenario.
The optimal values of $\tau$ are larger for datasets with a large number of clusters. In this scenario, values of 0.7 to 0.9
are preferred. For the other cases, the optimal value hovers around 0.5.
In order to recommend a unique value, we took the average among the optimal values of $\tau$ for each scenario.
This yielded $\tau = 0.6$.

To evaluate the sensitivity of the bandwidth parameter $\hat\sigma^2$ through the tuning parameter $s$ in \eqref{gausskern},
we evaluate the ARI criterion as above for every value of $s \in \{0.5,0.75,1,1.25,1.5,2\}$.
Following the above observations on the optimal values for $a$, $R$ and $\tau$, for this experiment, we
set $a=0.5$, $R=200$, and $\tau=0.6$.
The simulation results clearly show that the value of $s$ has almost no effect on the clustering configurations, nor the corresponding ARI values.
Therefore, we decided to fix $s=1$ in \eqref{gausskern}.

\section{Comparison performance between consensus DPP and PAM}\label{sec:performance}

As the choices for $a$, $R$, $\tau$ and $s$ have been fixed, we can now assess the performance of our clustering algorithm. For the evaluation, we use the procedure to simulate data described in the previous section.
As a reference, and for comparison purposes,
our consensus clustering methodology will not only be applied to clustering configurations generated with DPP, but also
to clustering configurations generated with the well-known
Partitioning Around Medoids (PAM) method \citep{Kaufmann1987}.
The PAM method is a classical partitioning technique for clustering that chooses the data point centers of the Voronoi cells
by simple random sampling.
As DPP selects data points centers based on diversity,
our goal here is to study how the quality of the clustering configurations depends on diversity at sampling centroids.

To sample points at random in the PAM procedure, we first sample a number $k$ of Voronoi cells uniformly at random from a finite
set of integers $\{1,\ldots, K_{\max}\}$. Then, we sampled uniformly at random $k$ points from the dataset.
Although, the sampling is uniform over the subset sizes, and over subsets of the same size,
this sampling technique is not really uniform; it favors very large and very small subsets over moderately sized subsets.
In fact, the probability of choosing a subset of $k$ points
$\{ x_{i_1}, \ldots, x_{i_k}\}$ following this sampling technique is
$\bigl( K_{\max}\binom{n}{k} \bigr)^{-1}.$
Despite this fact, we will refer to this sampling as uniform random sampling.

\begin{figure}[H]
    \centering

    \includegraphics[width=\textwidth,height=\textheight,keepaspectratio]{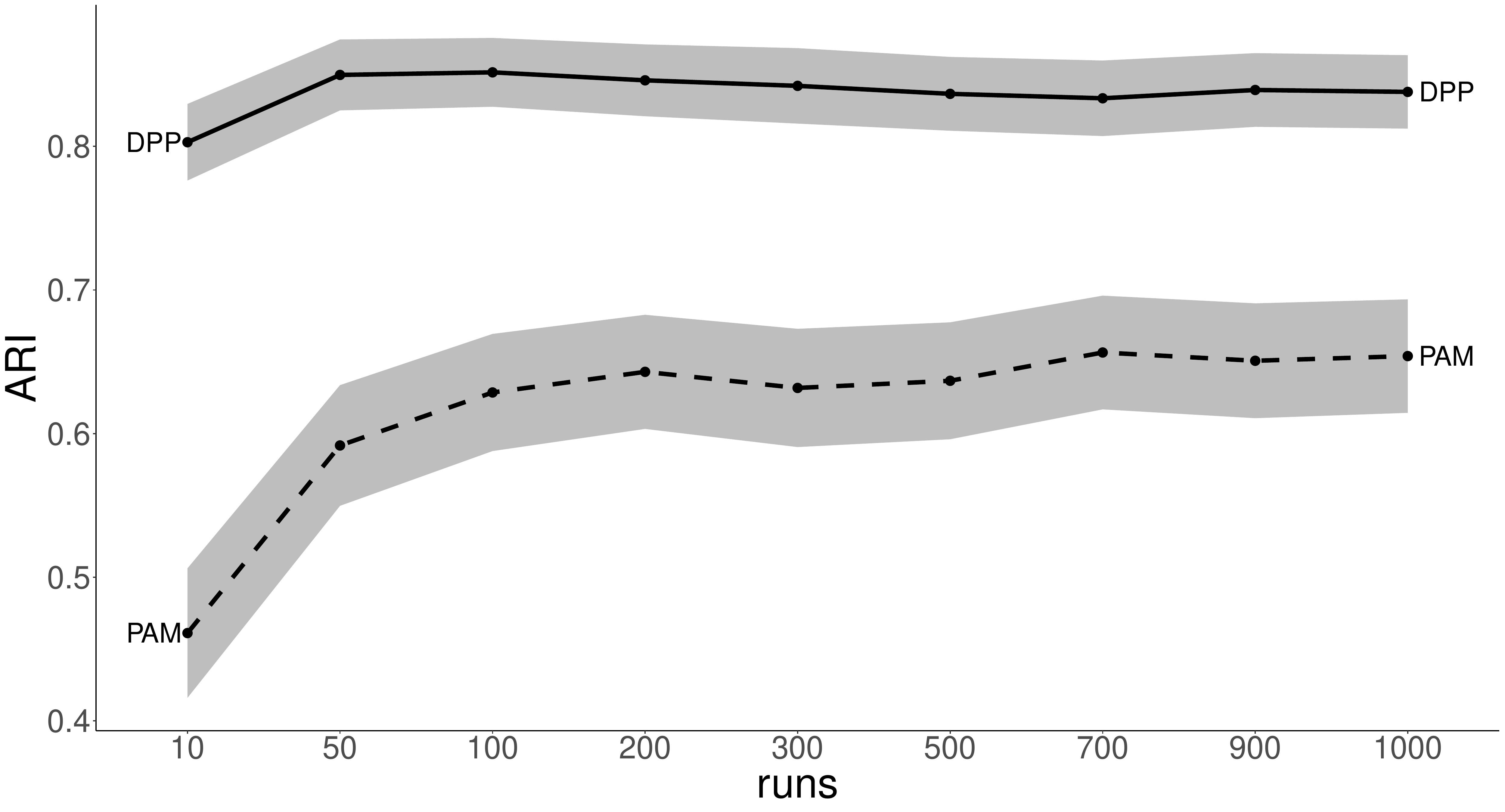}
    \caption{ARI mean trajectories for DPP and PAM as a function of the number of runs $R$.
      The regions encompassing the trajectories are the envelops of the corresponding 95\% confidence intervals
      associated with each mean ARI value of the run $R$.}
    \label{DPPvsPAM}
    \end{figure}

First, we compare the ARI trajectories of consensus DPP with those of PAM
as a function on the number of runs $R$.
Figure~\ref{DPPvsPAM} displays these trajectories considering all experimental scenarios
described in the previous section. The trajectories globally reflect the performance of each experimental condition involving the three levels of variables (low, medium and large) and the three levels of clusters (low, medium and large).
Observe that due to the diversity in the sampling of data points center, DPP has a jump-start like behavior, requiring very
few runs to achieve good clustering configurations. PAM, on the contrary, improves slowly its clustering configurations, and sometimes, even after 1000 runs, it never catches up with consensus DPP.

 We can see clear benefits of using DPP as a sampling method for the initial points needed to construct the Voronoi diagrams
 that define the clustering configurations.
 However, consensus DPP and PAM yield similar results when the variable dimensions are low.
 The dimension $p$ seems to play a crucial role in the potential of DPPs to sample with more diversity:
 there are more possibilities of distinguishing
 two vectors $x_i,x_j\in\mathds{R}^p$ in the transformed space $\mathcal{H}$
 when the dimension $p$ is already large, since the two vectors are more likely to be projected in very different
 places.
 In this case, DPP will sample these two points together more often than uniform random sampling,
 which will sample any pair of points with the same probability.
 When $p$ is small, the two vectors have less potential of being very different, and
 using DPP or PAM should give similar results.

\begin{figure}[H]
\centering

\subfloat[First simulated dataset]{%
\includegraphics[width=0.31\textwidth]{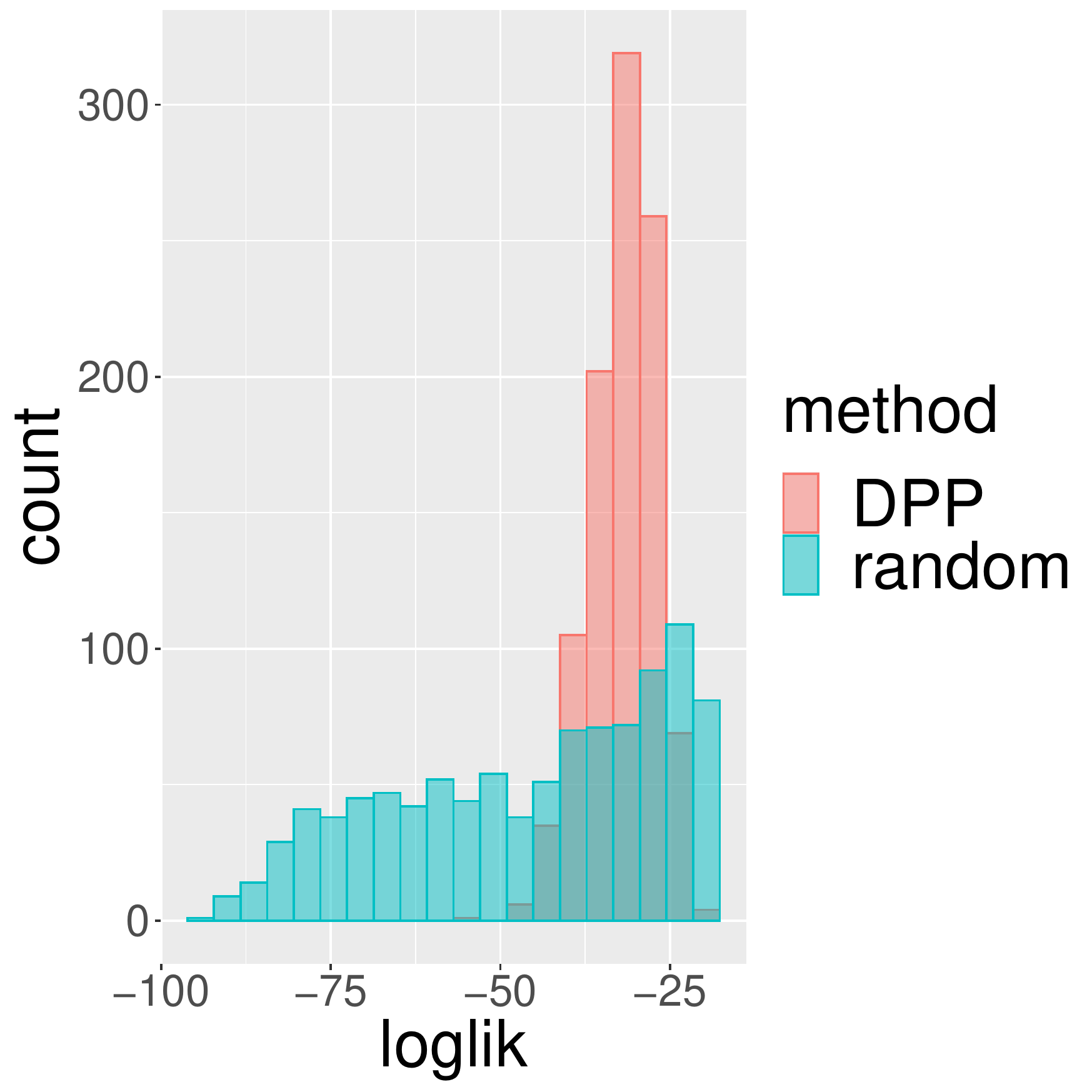}}
\quad
\subfloat[Second simulated dataset]{%
\includegraphics[width=0.31\textwidth]{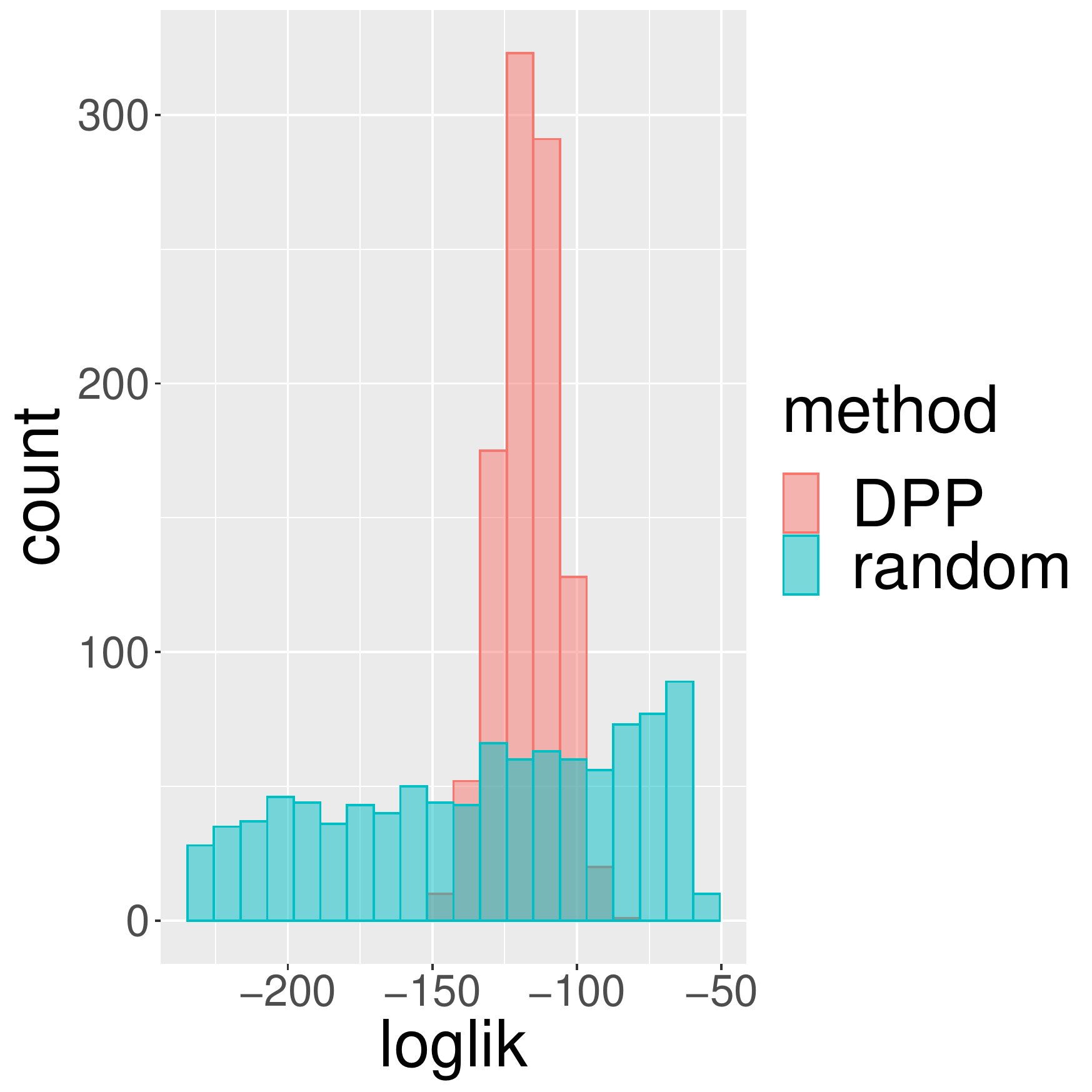}}
\quad
\subfloat[Third simulated dataset]{%
\includegraphics[width=0.31\textwidth]{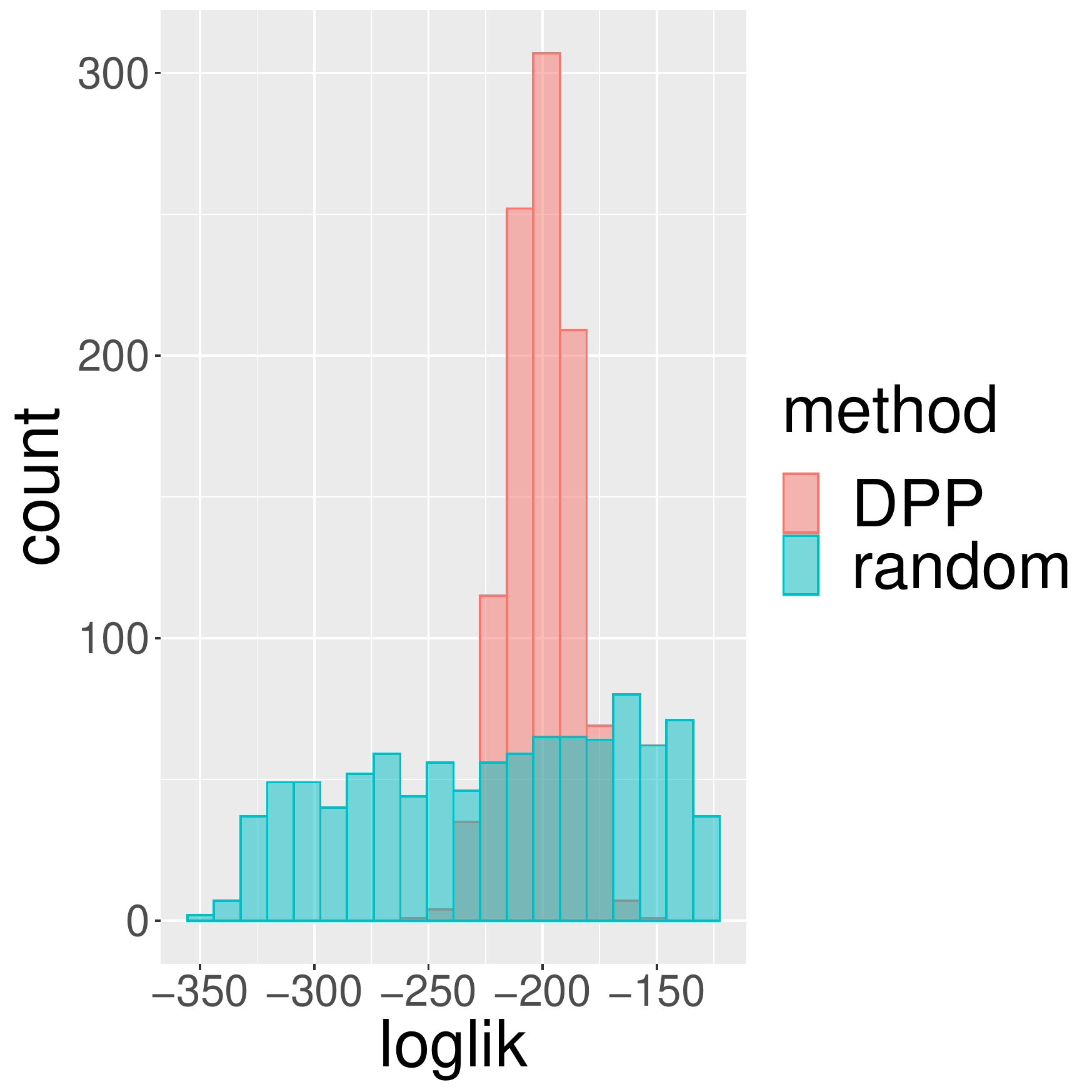}}
\caption{Typical histograms of the logarithm of the probability mass function (loglik), using DPP (light bars) and simple random sampling (dark bars), for three simulated datasets.}
\label{diversitysim}
\end{figure}

 To complement the differences between DPP and PAM, we show in Figure~\ref{diversitysim}
 typical histograms of the logarithm of the probability mass function of the DPP, given by \eqref{definition22},
 for 1000 sampled random subsets, using either the DPP sampling algorithm of \cite{BenHough2006} and \cite{Kulesza2012},
 or the uniform random sampling of PAM,
 for three simulated datasets selected among the 24 experimental scenarios.
 \setcounter{subfigure}{0}
The histograms clearly show that DPP selects random subsets with higher and less dispersed probability mass values (likelihood)
than uniform random sampling.
This explains the observed lower dispersion of the ARI values when sampling is performed with DPP.
The higher likelihood of the random subsets sampled by DPP confirms the higher diversity of those subsets.
Instead, subsets sampled as in PAM can be highly or poorly diverse: in fact, the associated histograms show very high dispersion in terms of diversity. DPP tends to select points that maintain a high level of diversity at each sampling,
proving to be more consistent and stable than uniform random sampling in terms of ensuring
the heterogeneity of the elements forming the subset.
\vspace{10pt}

Next, we fix $R=200$ as suggested by the study of the previous section, and compare the performance of consensus DPP
and PAM.
We have already noted the superiority of consensus DPP when looking at ARI as a measure of clustering quality.
But, this time we apply our mixed strategy presented in Section~\ref{sec:consensus:DPP}.
That is, we consider a sequence of thresholds $\theta_1, \theta_2, \dots, \theta_t$ that are above the inferior limit $\tau$.
We recall that adopting the strategy of considering a range of thresholds results in a collection
of consolidated clustering configurations for each datasets in each scenario.
For the choice of the optimal clustering configuration, we  use the kernel-based validation index KVI$_\mathcal{V}$
in \eqref{KRV}. To measure the goodness-of-fit of the optimal clustering configuration we use the ARI and RN measures
described earlier.
Table~\ref{table:comparison:resultsARI} displays the ARI means and standard deviations over all 24 scenarios,
while Table~\ref{table:comparison:resultsRN} displays the RN means and standard deviations over all 24 scenarios.

\begin{table}[H]
\centering
\begin{tabular}{cccccc}
\toprule
\multirow{2}{*}{\begin{tabular}[c]{@{}c@{}}Sample \\ size\end{tabular}} & \multirow{2}{*}{Method} & \multirow{2}{*}{\begin{tabular}[c]{@{}c@{}}Number of\\ clusters\end{tabular}} & \multicolumn{3}{c}{Number of variables} \\ \cmidrule{4-6}
                                                                        &                         &                                                                               & Low        & Medium        & Large       \\ \midrule
\multirow{4}{*}{$n=150$}                                                & DPP                     & \multirow{2}{*}{Low}                                                          & 0.95\,(0.07)        & 0.92\,(0.16)           & 0.90\,(0.15)        \\
                                                                        & PAM                     &                                                                               & 0.90\,(0.14)          & 0.93\,(0.13)             & 0.74\,(0.22)          \\ \cmidrule{3-6}
                                                                        & DPP                     & \multirow{2}{*}{Medium}                                                       & 0.92\,(0.09)        & 0.94\,(0.07)           & 0.85\,(0.10)        \\
                                                                        & PAM                     &                                                                               & 0.98\,(0.02)          & 0.83\,(0.14)             & 0.65\,(0.14)          \\ \midrule
\multirow{6}{*}{$n=500$}                                                & DPP                     & \multirow{2}{*}{Low}                                                          & 0.95\,(0.11)        & 0.95\,(0.10)           & 0.92\,(0.09)        \\
                                                                        & PAM                     &                                                                               & 0.77\,(0.22)          & 0.86\,(0.19)             & 0.83\,(0.18)          \\ \cmidrule{3-6}
                                                                        & DPP                     & \multirow{2}{*}{Medium}                                                       & 0.95\,(0.08)        & 0.98\,(0.03)           & 0.91\,(0.11)        \\
                                                                        & PAM                     &                                                                               & 0.96\,(0.08)          & 0.90\,(0.11)             & 0.71\,(0.34)          \\ \cmidrule{3-6}
                                                                        & DPP                     & \multirow{2}{*}{Large}                                                         & 0.96\,(0.05)        & 0.96\,(0.03)           & 0.98\,(0.02)        \\
                                                                        & PAM                     &                                                                               & 0.99\,(0.02)          & 0.87\,(0.10)             & 0.70\,(0.15)          \\ \midrule
\multirow{6}{*}{$n=1500$}                                               & DPP                     & \multirow{2}{*}{Low}                                                          & 0.91\,(0.08)        & 0.89\,(0.11)           & 0.90\,(0.10)        \\
                                                                        & PAM                     &                                                                               & 0.66\,(0.26)          & 0.74\,(0.18)             & 0.76\,(0.26)          \\ \cmidrule{3-6}
                                                                        & DPP                     & \multirow{2}{*}{Medium}                                                       & 0.96\,(0.04)        & 0.99\,(0.01)           & 0.88\,(0.15)        \\
                                                                        & PAM                     &                                                                               & 0.98\,(0.02)          & 0.99\,(0.01)             & 0.96\,(0.06)          \\ \cmidrule{3-6}
                                                                        & DPP                     & \multirow{2}{*}{Large}                                                         & 0.97\,(0.04)        & 0.96\,(0.05)           & 0.96\,(0.02)        \\
                                                                        & PAM                     &                                                                               & 0.93\,(0.12)          & 0.90\,(0.17)             & 0.69\,(0.35)          \\
\bottomrule
\end{tabular}
\caption{ARI means and standard deviations (within parentheses) over all 24 scenarios with consensus DPP and PAM.}
\label{table:comparison:resultsARI}
\end{table}

\begin{table}[H]
\centering
\begin{tabular}{cccccc}
\toprule
\multirow{2}{*}{\begin{tabular}[c]{@{}c@{}}Sample \\ size\end{tabular}} & \multirow{2}{*}{Method} & \multirow{2}{*}{\begin{tabular}[c]{@{}c@{}}Number of\\ clusters\end{tabular}} & \multicolumn{3}{c}{Number of variables} \\ \cmidrule{4-6}
                                                                        &                         &                                                                               & Low        & Medium        & Large       \\ \midrule
\multirow{4}{*}{$n=150$}                                                & DPP                     & \multirow{2}{*}{Low}                                                          & 0.03\,(0.06)        & 0.04\,(0.08)           & 0.04\,(0.06)        \\
                                                                        & PAM                     &                                                                               & 0.04\,(0.08)          & 0.03\,(0.08)             & 0.09\,(0.12)          \\ \cmidrule{3-6}
                                                                        & DPP                     & \multirow{2}{*}{Medium}                                                       & 0.03\,(0.05)        & 0.02\,(0.03)           & 0.04\,(0.04)        \\
                                                                        & PAM                     &                                                                               & 0.00\,(0.00)          & 0.08\,(0.08)             & 0.17\,(0.07)          \\ \midrule
\multirow{6}{*}{$n=500$}                                                & DPP                     & \multirow{2}{*}{Low}                                                          & 0.05\,(0.13)        & 0.04\,(0.13)           & 0.04\,(0.08)        \\
                                                                        & PAM                     &                                                                               & 0.23\,(0.21)          & 0.08\,(0.11)             & 0.13\,(0.17)          \\ \cmidrule{3-6}
                                                                        & DPP                     & \multirow{2}{*}{Medium}                                                       & 0.03\,(0.06)        & 0.02\,(0.05)           & 0.04\,(0.07)        \\
                                                                        & PAM                     &                                                                               & 0.02\,(0.06)          & 0.05\,(0.06)             & 0.16\,(0.20)          \\ \cmidrule{3-6}
                                                                        & DPP                     & \multirow{2}{*}{Large}                                                         & 0.02\,(0.02)        & 0.02\,(0.03)           & 0.01\,(0.01)        \\
                                                                        & PAM                     &                                                                               & 0.01\,(0.01)          & 0.06\,(0.06)             & 0.14\,(0.06)          \\ \midrule
\multirow{6}{*}{$n=1500$}                                               & DPP                     & \multirow{2}{*}{Low}                                                          & 0.11\,(0.13)        & 0.16\,(0.21)           & 0.18\,(0.22)        \\
                                                                        & PAM                     &                                                                               & 0.50\,(0.40)          & 0.37\,(0.28)             & 0.44\,(0.50)          \\ \cmidrule{3-6}
                                                                        & DPP                     & \multirow{2}{*}{Medium}                                                       & 0.02\,(0.04)        & 0.00\,(0.00)           & 0.16\,(0.21)        \\
                                                                        & PAM                     &                                                                               & 0.01\,(0.03)          & 0.00\,(0.00)             & 0.03\,(0.08)          \\ \cmidrule{3-6}
                                                                        & DPP                     & \multirow{2}{*}{Large}                                                         & 0.02\,(0.02)        & 0.02\,(0.03)           & 0.01\,(0.03)        \\
                                                                        & PAM                     &                                                                               & 0.03\,(0.06)          & 0.04\,(0.07)             & 0.14\,(0.16)          \\
\bottomrule
\end{tabular}
\caption{RN means and standard deviations (within parentheses) over all 24 scenarios with consensus DPP and PAM.}
\label{table:comparison:resultsRN}
\end{table}

      As noted earlier, there is a clear advantage of using DPP over uniform random sampling for the initial points needed to construct the Voronoi diagrams.
      This is particularly true when the number of variables is moderate to high.
      When the number of variables is low, the results yielded by DPP and PAM are similar.
      Another interesting advantage that we observe is that
      sampling with DPP
      contributes to reducing the dispersion of the ARI scores,
      and then produces more stable optimal clustering configurations.
      Turning now to the RN means of Table~\ref{table:comparison:resultsRN}, the same conclusion applies:
      sampling with DPP yields better results. The optimal clustering configurations yielded by consensus DPP
      are associated with estimated number of clusters closer to the true number of clusters than those yielded by PAM.
      Observe as well, that DPP contributes to the reduction of the variability of the RN values for almost all the cases,
      as it was already the case with ARI.

\section{Application to real data}\label{sec:real1}

In this section we proceed to evaluate the performance of consensus DPP versus PAM on real datasets. The datasets
were obtained from the \emph{UCI Machine Learning Repository} \citep{Dua2019} and \emph{OpenML} website \citep{OpenML2013}, two well known databases in the Machine Learning community for clustering and classification problems. Table~\ref{tauchoicereal} shows the selected real datasets and some of their features: $n=$ number of observations, $K=$ number of clusters,
$p=$ number of variables (i.e., data dimension).

\begin{table}[H]

\begin{center}
\begin{tabular}{lrrr}
  \hline
  Dataset &$n$ & $K$& $p$  \\\hline
  Iris & 150 & 3&4  \\
  OliveOil & 572 & 9 & 8  \\
  Ecoli & 327 & 5 & 7 \\
  Bank & 1372 & 2 & 4 \\
  Colposcopy & 287 & 3 & 62  \\
  Forest & 198 & 4 & 27  \\
  Breast & 569 & 2 & 30  \\
  Synthetic & 600 & 6 & 60  \\
   Lung cancer & 181 & 2 & 12533 \\
  Yeast Cycle & 384 & 5 & 17  \\
  \hline
\end{tabular}
\caption{Selected datasets from the UCI and openML repositories}
\label{tauchoicereal}
\end{center}
\end{table}
Following the recommendations in \citep{Bicego2016,Xuan2013}, and
due to its strongly unbalanced nature,
the Ecoli dataset was transformed using the Box-Cox transformation procedure. Moreover, the original dataset contains $n=336$ observations with $K=8$ clusters, but two clusters have only 2 observations, and a third cluster has only 5 observations. These clusters were removed from the data.
The Breast and Lung Cancer datasets were also transformed using the Box-Cox transformation.
We note that transforming the data is a common procedure for DNA microarray data \citep{Thygesen2004}.
The Bank dataset has only two clusters, even though it contains $n=1372$ observations.
So using $\sqrt{n}\approx 37$ as a minimal cluster size in the cluster merging stage of
the consensus procedure is not optimal.
We note that our experiments to select the appropriate parameters for consensus DPP
hinted at larger values of the power $a$ when the number of clusters is small.
Hence, for these data, we used $n^{2/3}\approx 124$ as the minimal cluster size.

For each dataset, we performed $R=200$ runs of consensus DPP. The procedure was
repeated ten times. For the choice of the optimal clustering configuration,
we use the kernel-based validation index KVI$_\mathcal{V}$ defined in \eqref{KRV}.
To measure the goodness-of-fit of the optimal clustering configuration we
use the ARI and RN measures.
We compare the consensus DPP results to those of two traditional clustering algorithms: PAM and $k$-means.
Our goal is to show the advantages of the DPP diversity at sampling centroids
on the quality of clustering configurations.

The PAM algorithm was already used and mentioned in
the study with the simulated datasets of Section~\ref{sec:performance}.
The $k$-means algorithm was proposed by Stuart Lloyd in 1957, and later published in \citep{Lloyd19822}.
It starts with an initial set of $k$ means, representing $k$ clusters.
It assigns each observation to the corresponding Voronoi cell or cluster given by the corresponding
closer mean among the $k$ means. Once all observations are assigned, the mean vectors of all
Voronoi cells are updated, and the process is repeated
until there is no change in the means.
However, as argued by \cite{Celebi2013}, the popular methods for choosing
the initial set of $k$ means, such as Forgy \citep{Forgy1965}, Random Partition \citep{Pena1999} and Maximin methods \citep{Gonzalez1985,Katsavounidis1994}, result often in cluster configurations with a low clustering quality.
For that reason we decided to work with the $k$-means$++$ algorithm of \cite{Arthur2007},
a popular choice mentioned by several authors \citep{Capo2017,Franti2019} that avoids the poor quality results of
the traditional methods for choosing the initial means. It is based on a simple probabilistic technique.
  For the consensus clustering with $k$-means, we proceed as follows: we first sample a number $k$ of Voronoi cells uniformly at random from a finite
  set of integers $\{1,\ldots, K_{\max}\}$.
  Then we run $k$-means$++$ to obtain an initial set of $k$ means. This step consists of
  (i) selecting the first center at random from the dataset $\mathcal{S}$; and
  (ii) repeating the following two steps until a subset of $k$ centers has been sampled:
  (a) for each $x \in\mathcal{S}$, we compute $dc2(x)$ the square of the Euclidean distance between $x$
  and the closest center among those already sampled;
  (b) a new center is sampled with probability $dc2(x)/\sum_{x'\in\mathcal{S}}dc2(x')$.
  Once the $k$ centers have been chosen, we proceed as in the standard $k$-means algorithm described above.
  We repeat the selection of $k$ initial means $R=200$ times, just as we do with DPP and PAM,
  to afterwards apply our consensus clustering methodology to the clustering configurations.
  As we did with consensus DPP, the optimal cluster configurations from PAM and $k$-means were chosen
  using the kernel-based validation index KVI$_\mathcal{V}$ criterion defined in \eqref{KRV}.
  The whole procedure was repeated ten times.

Table~\ref{ARI-DPPvsPAM} displays the ARI and RN means
and standard deviations obtained by applying consensus DPP, PAM and $k$-means consensus clustering
to the datasets of Table~\ref{tauchoicereal}.

\begin{table}[H]
\centering
\begin{tabular}{cclll} \\ \toprule
  Dataset                                           & Measure & DPP          & PAM          & $k$-means    \\ \midrule
\multicolumn{1}{c}{\multirow{2}{*}{Iris}}         & ARI & 0.91\,(0.03) & 0.83\,(0.09) & 0.66\,(0.05) \\
                                                  & RN  & 0.03\,(0.07) & 0.06\,(0.08) & 0.02\,(0.05) \\ \midrule
\multicolumn{1}{c}{\multirow{2}{*}{OliveOil}}     & ARI & 0.72\,(0.06) & 0.60\,(0.11) & 0.68\,(0.08) \\
                                                  & RN  & 0.12\,(0.05) & 0.10\,(0.07) & 0.11\,(0.03) \\  \midrule
\multicolumn{1}{c}{\multirow{2}{*}{Ecoli}}        & ARI & 0.76\,(0.02) & 0.66\,(0.09) & 0.71\,(0.07) \\
                                                  & RN  & 0.05\,(0.06) & 0.14\,(0.08) & 0.04\,(0.05) \\  \midrule
\multicolumn{1}{c}{\multirow{2}{*}{Bank}}         & ARI & 0.66\,(0.09) & 0.53\,(0.19) & 0.50\,(0.10) \\
                                                  & RN  & 0.13\,(0.12) & 0.25\,(0.19) & 0.24\,(0.15) \\  \midrule
\multicolumn{1}{c}{\multirow{2}{*}{Colposcopy}}   & ARI & 0.44\,(0.09) & 0.42\,(0.14) & 0.35\,(0.11) \\
                                                  & RN  & 0.21\,(0.09) & 0.15\,(0.10) & 0.16\,(0.14) \\ \midrule
\multicolumn{1}{c}{\multirow{2}{*}{Forest}}       & ARI & 0.86\,(0.05) & 0.70\,(0.01) & 0.72\,(0.22) \\
                                                  & RN  & 0.01\,(0.04) & 0.13\,(0.00) & 0.08\,(0.12) \\ \midrule
\multicolumn{1}{c}{\multirow{2}{*}{Breast}}       & ARI & 0.61\,(0.05) & 0.50\,(0.13) & 0.61\,(0.13) \\
                                                  & RN  & 0.09\,(0.12) & 0.13\,(0.12) & 0.07\,(0.11) \\ \midrule
\multicolumn{1}{c}{\multirow{2}{*}{Synthetic}}    & ARI & 0.69\,(0.02) & 0.66\,(0.04) & 0.64\,(0.01) \\
                                                  & RN  & 0.09\,(0.10) & 0.16\,(0.08) & 0.13\,(0.07) \\ \midrule
\multicolumn{1}{c}{\multirow{2}{*}{Lung cancer}}  & ARI & 0.89\,(0.14) & 0.84\,(0.34) & 0.65\,(0.42) \\
                                                  & RN  & 0.09\,(0.12) & 0.00\,(0.00) & 0.02\,(0.07) \\ \midrule
\multicolumn{1}{c}{\multirow{2}{*}{Yeast Cycle}}  & ARI & 0.47\,(0.004) & 0.47\,(0.03) & 0.44\,(0.05) \\
                                                  & RN  & 0.08\,(0.06) & 0.10\,(0.04) & 0.03\,(0.05) \\
\bottomrule
\end{tabular}
\caption{ARI and RN means and standard deviations (within parentheses)
  over ten runs associated with consensus DPP, PAM, and $k$-means.}
\label{ARI-DPPvsPAM}
\end{table}
We observe that consensus DPP yields higher
ARI values than the two other methods,
and contributes to reducing the variability of this measure, as well.
That is, consensus DPP produces more stable and better clustering configurations.
The RN results are more balanced in the sense that there is no major difference
between the three methods. This means that
all three methods hinted at reasonable number of clusters, but not all got
good clustering configurations.

\begin{figure}[H]
\centering

\subfloat[Iris dataset]{%
\includegraphics[width=0.45\textwidth]{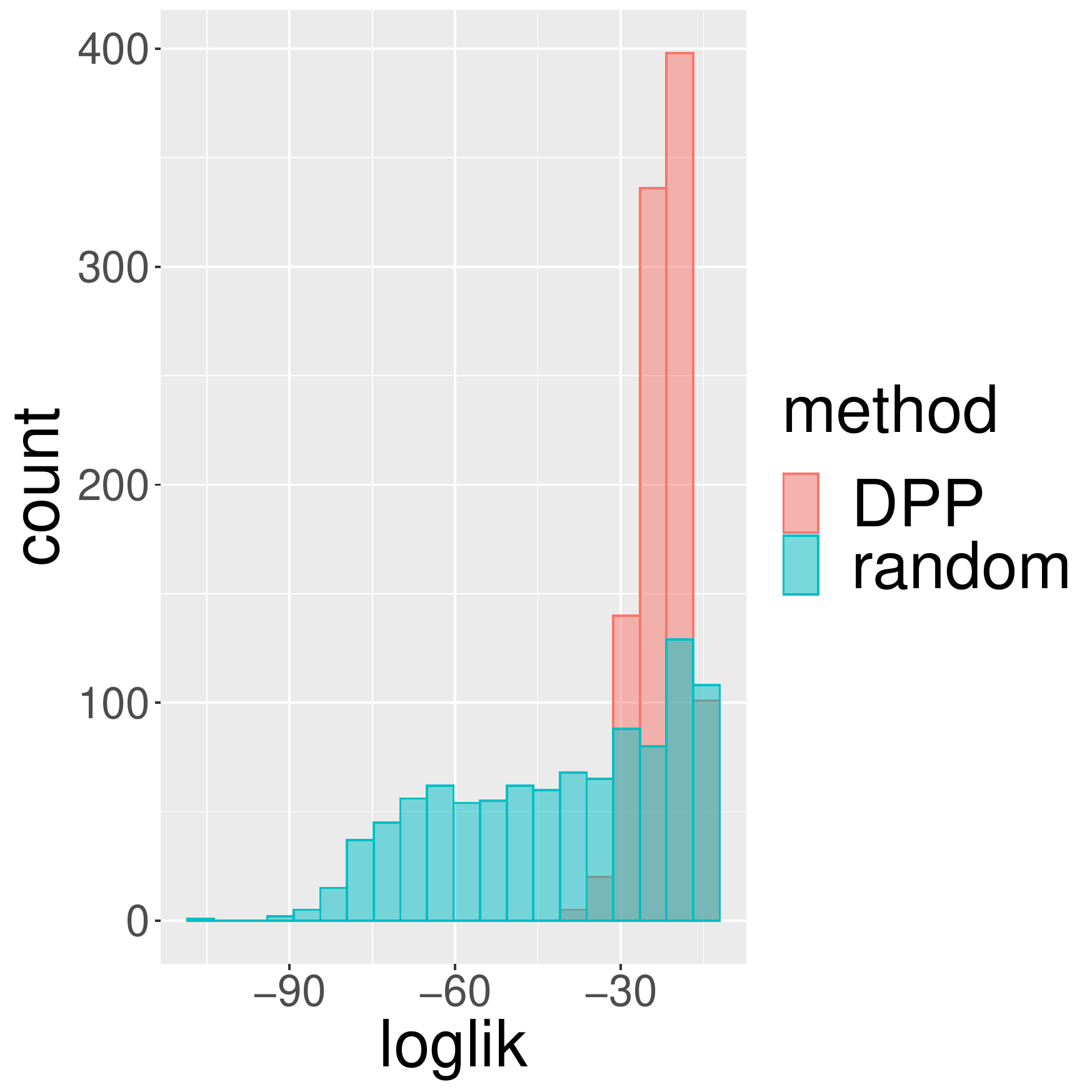}}
\quad
\subfloat[Synthetic dataset]{%
\includegraphics[width=0.45\textwidth]{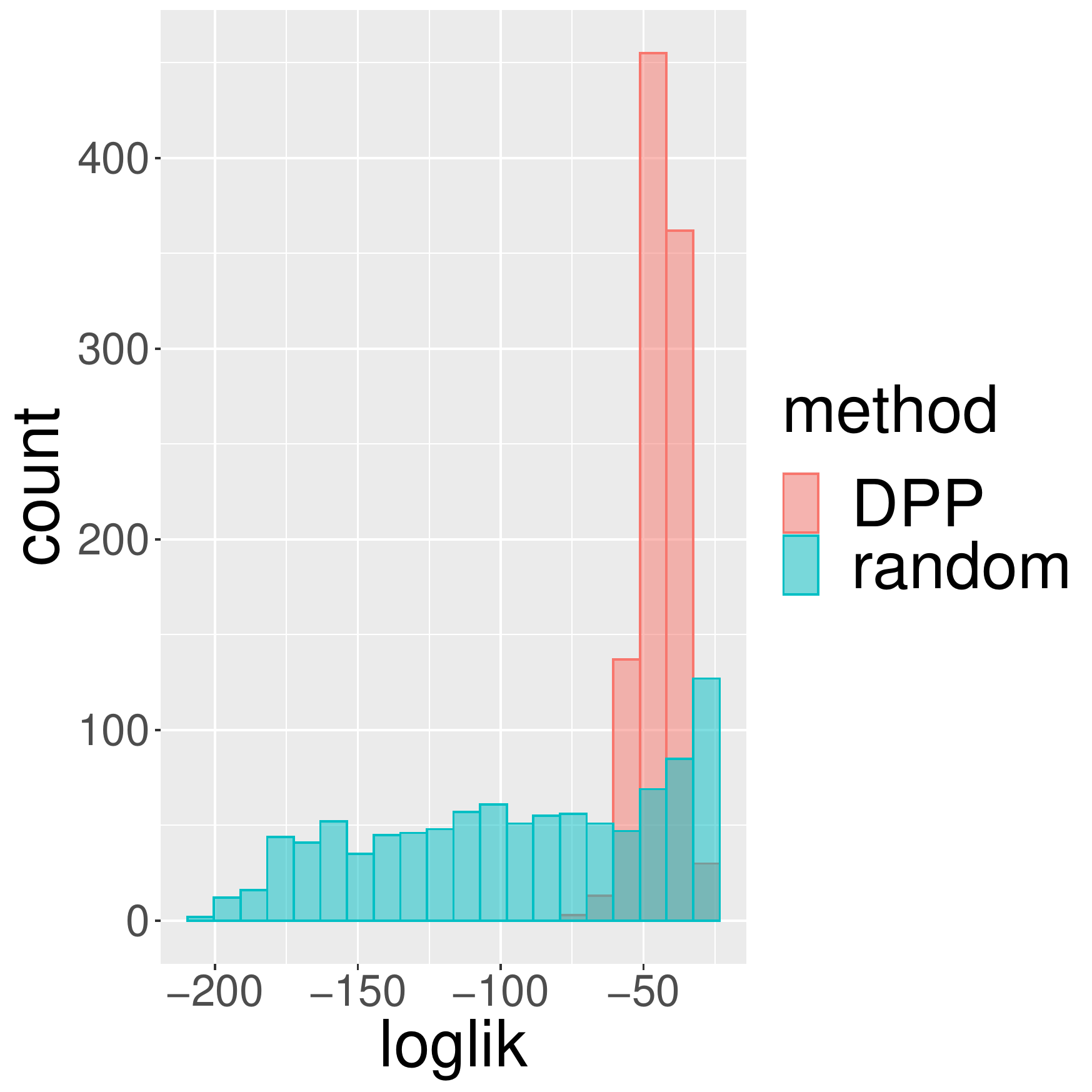}}
\caption{Typical histograms of the logarithm of the probability mass function (loglik) using DPP and uniform random sampling, for two real datasets.}
\label{diversityreal}
\end{figure}

As we also did for the simulated data in Section \ref{sec:performance},
we present in Figure~\ref{diversityreal} two
typical histograms of the logarithm of the probability mass function of the DPP, given by \eqref{definition22}.
The datasets in the figure are {\it Iris} and {\it Synthetic} (see Table~\ref{tauchoicereal}).

 \setcounter{subfigure}{0}

As with simulated datasets, we observe that while subsets sampled at random as in PAM result in histograms with a very high dispersion in terms of diversity, DPP tends to select points that maintain a high level of diversity in each sample.
DPP seems to be more consistent and stable than uniform random sampling in
ensuring the heterogeneity of the elements forming the subsets.

\section{Conclusions}\label{sec:conclusions1}

  We explored the potential of determinantal point processes as a sampling method for initializing each run of a consensus clustering algorithm.
  As a probabilistic model of repulsion, it favors diversity within subsets of points.
  This is in contrast to uniform random sampling, which gives to every point an equal probability of being selected.
  Extended simulations showed that, when compared to uniform random sampling,
  the use of DPPs to generate initial subsets of points results in final clustering configurations with higher and less dispersed quality scores.
  Applications to real datasets confirm these conclusions draw from simulations.

      By using DPPs to generate center point subsets for clustering, the consensus clustering does not require a large number of sampled partitions to ensure a high goodness-of-fit score (e.g., ARI) in the final clustering configurations.
    In fact, a moderate number of ensemble partitions of about 100 or 200 is sufficient.
    In contrast, uniform random sampling generally requires a larger number of sampled partitions to reach ARI mean values comparable to determinantal consensus clustering. For the choice of the final clustering configuration among several candidates, the kernel-based validation index of \cite{Fan2010} has proven to be a good option, outperforming other indexes.

    Selecting an appropriate threshold during the merging procedure of the determinantal consensus algorithm is essential.
Our simulations show that a good strategy consists in
choosing a small subset of diverse thresholds among all the
    possible threshold values given by the observed consensus indexes.
    The main advantage of this strategy is to speed up the computations, while preserving the properties associated with
    keeping all threshold values  from the set of all different observed consensus indexes \citep{Murua2014}.
    Retaining thresholds above 0.6 was adopted as a general choice.

 A variety of interesting questions remain for future research:
  (i) To extend the determinantal consensus clustering to datasets with both continuous and categorical variables, inducing the choice of a proper measure of distance, necessary for the construction of the kernel matrix.
  In general, for continuous random variables, the Euclidean or Mahalanobis-like distances perform well.
  For categorical data, Lin's pairwise similarity measure \citep{lin-1998} is an attractive alternative to the usual Hamming distance.
  (ii) To study the effect of multivariate outliers on the mean and dispersion of quality scores
  of clustering configurations yielded by the determinantal consensus clustering.
(iii) To adapt the determinantal consensus clustering to the case of very large datasets.
  The bottleneck of the method is the eigendecomposition of the kernel matrix. This is a central step for obtaining an initial random subset of points with the determinantal point process. The computational complexity of the eigendecomposition of a $n\times n$ symmetric matrix is $O\left(n^3\right)$. As $n$ grows larger,
  the computation of the matrix spectral decomposition
  becomes expensive.
  We explore approximative ways to overcome this challenge with sparse matrix approximation to the kernel matrix.
  This is the topic we cover in a sequel paper on determinantal consensus clustering.

\bibliographystyle{plain}
\bibliography{References1}  

\begin{thebibliography}{10}

\bibitem{Ankerst1999}
Mihael Ankerst, Markus~M Breunig, Hans-Peter Kriegel, and J{\"o}rg Sander.
\newblock Optics: ordering points to identify the clustering structure.
\newblock {\em ACM Sigmod record}, 28(2):49--60, 1999.

\bibitem{Ao2004}
Sio-Iong Ao, Kevin Yip, Michael Ng, David Cheung, Pui Fong, Ian Melhado, and
  Pak Sham.
\newblock Clustag: Hierarchical clustering and graph methods for selecting tag
  snps.
\newblock {\em Bioinformatics (Oxford, England)}, 21:1735--6, 05 2005.

\bibitem{Arthur2007}
David Arthur and Sergei Vassilvitskii.
\newblock K-means++: The advantages of careful seeding.
\newblock In {\em Proceedings of the Eighteenth Annual ACM-SIAM Symposium on
  Discrete Algorithms}, {SODA '07}, pages 1027--1035, USA, 2007. Society for
  Industrial and Applied Mathematics.

\bibitem{Aurenhammer1991}
Franz Aurenhammer.
\newblock Voronoi diagrams\&mdash;a survey of a fundamental geometric data
  structure.
\newblock {\em ACM Comput. Surv.}, 23(3):345--405, September 1991.

\bibitem{Banfield1993}
Jeffrey~D. Banfield and Adrian~E. Raftery.
\newblock Model-based {G}aussian and non-{G}aussian clustering.
\newblock {\em Biometrics}, 49(3):803--821, 1993.

\bibitem{BenHough2006}
J.~Ben~Hough, Manjunath Krishnapur, Yuval Peres, and B\'alint Vir\'ag.
\newblock Determinantal processes and independence.
\newblock {\em Probability Surveys [electronic only]}, 3:206--229, 2006.

\bibitem{Bicego2016}
Manuele Bicego and Sisto Baldo.
\newblock Properties of the box--cox transformation for pattern classification.
\newblock {\em Neurocomputing}, 218:390--400, 2016.

\bibitem{Bien2011}
Jacob Bien and Robert Tibshirani.
\newblock Hierarchical clustering with prototypes via minimax linkage.
\newblock {\em Journal of the American Statistical Association},
  106:1075--1084, 09 2011.

\bibitem{Blatt1996}
Marcelo Blatt, Shai Wiseman, and Eytan Domany.
\newblock Superparamagnetic clustering of data.
\newblock {\em Phys. Rev. Lett.}, 76:3251--3254, Apr 1996.

\bibitem{Blatt1997}
Marcelo Blatt, Shai Wiseman, and Eytan Domany.
\newblock Data clustering using a model granular magnet.
\newblock {\em Neural Comput.}, 9(8):1805--1842, November 1997.

\bibitem{Borodin2000}
A.~{Borodin} and G.~{Olshanski}.
\newblock {Distributions on Partitions, Point Processes, and the Hypergeometric
  Kernel}.
\newblock {\em Communications in Mathematical Physics}, 211:335--358, 2000.

\bibitem{Borodin2003}
Alexei Borodin and Alexander Soshnikov.
\newblock Janossy densities determinantal ensembles.
\newblock {\em Journal of Statistical Physics}, 113:595--610, 01 2003.

\bibitem{Capo2017}
Marco Cap{\'o}, Aritz P{\'e}rez, and Jose~A Lozano.
\newblock An efficient approximation to the k-means clustering for massive
  data.
\newblock {\em Knowledge-Based Systems}, 117:56--69, 2017.

\bibitem{Celebi2013}
M.~Emre Celebi, Hassan~A. Kingravi, and Patricio~A. Vela.
\newblock A comparative study of efficient initialization methods for the
  k-means clustering algorithm.
\newblock {\em Expert Systems with Applications}, 40(1):200 -- 210, 2013.

\bibitem{chaudhuri2017mean}
Arin Chaudhuri, Deovrat Kakde, Carol Sadek, Laura Gonzalez, and Seunghyun Kong.
\newblock The mean and median criteria for kernel bandwidth selection for
  support vector data description.
\newblock In {\em 2017 IEEE International Conference on Data Mining Workshops
  (ICDMW)}, pages 842--849. IEEE, 2017.

\bibitem{Chen-et-al-2002}
G.~Chen, S.~A. Jaradat, N.~Banerjee, T.~S. Tanaka, M.~S.~H. Ko, and M.~Q.
  Zhang.
\newblock Evaluation and comparison of clustering algorithms in analyzing es
  cell gene expression data.
\newblock {\em Statistica Sinica}, pages 241--262, 2002.

\bibitem{Daley2003}
D.~Daley and D.~Vere~Jones.
\newblock {\em An introduction to the theory of point processes. Volume I:
  Elementary theory and methods}, volume~1.
\newblock Springer, 2 edition, 01 2003.

\bibitem{Dua2019}
Dheeru Dua and Casey Graff.
\newblock {UCI} machine learning repository, 2017.

\bibitem{Ester1996}
Martin Ester, Hans-Peter Kriegel, J{\"o}rg Sander, Xiaowei Xu, et~al.
\newblock A density-based algorithm for discovering clusters in large spatial
  databases with noise.
\newblock In {\em Kdd}, volume~96, pages 226--231, 1996.

\bibitem{Fan2010}
Zizhu Fan, Xiangang Jiang, Baogen Xu, and Zhaofeng Jiang.
\newblock An automatic index validity for clustering.
\newblock In Ying Tan, Yuhui Shi, and Kay~Chen Tan, editors, {\em Advances in
  Swarm Intelligence}, pages 359--366, Berlin, Heidelberg, 2010. Springer
  Berlin Heidelberg.

\bibitem{Fausshauer2011}
Gregory Fasshauer.
\newblock Positive definite kernels: Past, present and future.
\newblock {\em Dolomite Res. Notes Approx.}, 4, 01 2011.

\bibitem{Florek1951}
Kazimierz Florek, Jan {\L}ukaszewicz, Julian Perkal, Hugo Steinhaus, and Stefan
  Zubrzycki.
\newblock Sur la liaison et la division des points d'un ensemble fini.
\newblock In {\em Colloquium mathematicum}, volume~2, pages 282--285, 1951.

\bibitem{Forgy1965}
Edward~W Forgy.
\newblock Cluster analysis of multivariate data: efficiency versus
  interpretability of classifications.
\newblock {\em biometrics}, 21:768--769, 1965.

\bibitem{Fraley1998}
Chris Fraley and Adrian~E. Raftery.
\newblock How many clusters? which clustering method? answers via model-based
  cluster analysis.
\newblock {\em Computer Journal}, 41:578--588, 1998.

\bibitem{Franti2019}
Pasi Fr{\"a}nti and Sami Sieranoja.
\newblock How much can k-means be improved by using better initialization and
  repeats?
\newblock {\em Pattern Recognition}, 93:95--112, 2019.

\bibitem{Girolami2002}
M.~Girolami.
\newblock Mercer kernel-based clustering in feature space.
\newblock {\em IEEE Transactions on Neural Networks}, 13(3):780--784, May 2002.

\bibitem{Gonzalez1985}
Teofilo~F Gonzalez.
\newblock Clustering to minimize the maximum intercluster distance.
\newblock {\em Theoretical computer science}, 38:293--306, 1985.

\bibitem{Hafiz2014}
R.~{Hafiz Affandi}, E.~B. {Fox}, R.~P. {Adams}, and B.~{Taskar}.
\newblock {Learning the Parameters of Determinantal Point Process Kernels}.
\newblock {\em ArXiv e-prints}, February 2014.

\bibitem{Hafiz2013}
R.~{Hafiz Affandi}, E.~B. {Fox}, and B.~{Taskar}.
\newblock {Approximate Inference in Continuous Determinantal Point Processes}.
\newblock {\em ArXiv e-prints}, November 2013.

\bibitem{Han2011}
Jiawei Han, Micheline Kamber, and Jian Pei.
\newblock {\em Data Mining: Concepts and Techniques}.
\newblock Morgan Kaufmann Publishers Inc., San Francisco, CA, USA, 3rd edition,
  2011.

\bibitem{Hinneburg1999}
Alexander Hinneburg and Daniel~A Keim.
\newblock Optimal grid-clustering: Towards breaking the curse of dimensionality
  in high-dimensional clustering.
\newblock In {\em 25th International Conference on Very Large Databases}, pages
  506--517, 1999.

\bibitem{Horn20122}
Roger~A. Horn and Charles~R. Johnson.
\newblock {\em Matrix Analysis}.
\newblock Cambridge University Press, USA, 2nd edition, 2012.

\bibitem{Howley2006}
Tom Howley and Michael~G. Madden.
\newblock An evolutionary approach to automatic kernel construction.
\newblock In Stefanos Kollias, Andreas Stafylopatis, W{\l}odzis{\l}aw Duch, and
  Erkki Oja, editors, {\em Artificial Neural Networks -- ICANN 2006}, pages
  417--426, Berlin, Heidelberg, 2006. Springer Berlin Heidelberg.

\bibitem{Hubert1985}
Lawrence Hubert and Phipps Arabie.
\newblock Comparing partitions.
\newblock {\em Journal of Classification}, 2(1):193--218, Dec 1985.

\bibitem{Jain1988}
Anil Jain and Richard Dubes.
\newblock {\em Algorithms for Clustering Data}.
\newblock Prentice-Hall, Inc., Upper Saddle River, NJ, USA, 1988.

\bibitem{Kang20132}
Byungkon Kang.
\newblock {Fast Determinantal Point Process Sampling with Application to
  Clustering}.
\newblock In C.J.C. Burges, L.~Bottou, M.~Welling, Z.~Ghahramani, and K.Q.
  Weinberger, editors, {\em Advances in Neural Information Processing Systems
  26}, pages 2319--2327. Curran Associates, Inc., 2013.

\bibitem{Katsavounidis1994}
Ioannis Katsavounidis, C-C~Jay Kuo, and Zhen Zhang.
\newblock A new initialization technique for generalized lloyd iteration.
\newblock {\em IEEE Signal processing letters}, 1(10):144--146, 1994.

\bibitem{Kaufmann1987}
Leonard Kaufmann and Peter Rousseeuw.
\newblock Clustering by means of medoids.
\newblock {\em Data Analysis based on the L1-Norm and Related Methods}, pages
  405--416, 01 1987.

\bibitem{Kulesza2012}
A.~{Kulesza} and B.~{Taskar}.
\newblock {Determinantal point processes for machine learning}.
\newblock {\em ArXiv e-prints}, July 2012.

\bibitem{Lanckriet2004}
Gert R.~G. Lanckriet, Nello Cristianini, Peter Bartlett, Laurent~El Ghaoui, and
  Michael~I. Jordan.
\newblock Learning the kernel matrix with semidefinite programming.
\newblock {\em J. Mach. Learn. Res.}, 5:27--72, December 2004.

\bibitem{Lavancier2015}
Fr{\'e}d{\'e}ric Lavancier, Jesper M{\o}ller, and Ege Rubak.
\newblock Determinantal point process models and statistical inference.
\newblock {\em Journal of the Royal Statistical Society: Series B (Statistical
  Methodology)}, 77(4):853--877, 2015.

\bibitem{lin-1998}
D.~Lin.
\newblock An information-theoretic definition of similarity.
\newblock In {\em Proceedings of the 15th International Conference on Machine
  Learning, Morgan Kaufmann, San Francisco, CA}, pages 296--304, 1998.

\bibitem{Lloyd19822}
Stuart~P. Lloyd.
\newblock Least squares quantization in pcm.
\newblock {\em IEEE Trans. Inf. Theory}, 28:129--136, 1982.

\bibitem{Macchi1975}
Odile Macchi.
\newblock {The Coincidence Approach to Stochastic Point Processes}.
\newblock {\em Advances in Applied Probability}, 7(1):83--122, 1975.

\bibitem{Melnykov2012}
Volodymyr Melnykov, Wei-Chen Chen, and Ranjan Maitra.
\newblock Mixsim: An r package for simulating data to study performance of
  clustering algorithms.
\newblock {\em Journal of Statistical Software, Articles}, 51(12):1--25, 2012.

\bibitem{Monti2003}
Stefano Monti, Pablo Tamayo, Jill Mesirov, and Todd Golub.
\newblock Consensus clustering: A resampling-based method for class discovery
  and visualization of gene expression microarray data.
\newblock {\em Machine Learning}, 52(1):91--118, 2003.

\bibitem{Murua2018}
Johanna Mu{\~n}oz and Alejandro Murua.
\newblock Building cancer prognosis systems with survival function clusters.
\newblock {\em Statistical Analysis and Data Mining: The ASA Data Science
  Journal}, 11(3):98--110, 2018.

\bibitem{Murua2008}
Alejandro Murua, Larissa Stanberry, and Werner Stuetzle.
\newblock On potts model clustering, kernel k-means, and density estimation.
\newblock {\em Journal of Computational and Graphical Statistics},
  17(3):629--658, 2008.

\bibitem{Murua2014}
Alejandro Murua and Nicolas Wicker.
\newblock {The Conditional-Potts Clustering Model}.
\newblock {\em Journal of Computational and Graphical Statistics},
  23(3):717--739, 2014.

\bibitem{Okabe2000}
Atsuyuki Okabe, Barry Boots, Kokichi Sugihara, and Sung~Nok Chiu.
\newblock {\em Spatial Tessellations: Concepts and Applications of {V}oronoi
  Diagrams}.
\newblock Series in Probability and Statistics. John Wiley and Sons, Inc., 2nd
  ed. edition, 2000.

\bibitem{Pena1999}
Jos{\'e}~M Pena, Jose~Antonio Lozano, and Pedro Larranaga.
\newblock An empirical comparison of four initialization methods for the
  k-means algorithm.
\newblock {\em Pattern recognition letters}, 20(10):1027--1040, 1999.

\bibitem{Rand1971}
William~M. Rand.
\newblock Objective criteria for the evaluation of clustering methods.
\newblock {\em Journal of the American Statistical Association},
  66(336):846--850, 1971.

\bibitem{Schlkopf2004}
Bernhard. Sch\"{o}lkopf, Koji. Tsuda, and Jean-Philippe. Vert.
\newblock {\em Kernel methods in computational biology}.
\newblock MIT Press, Cambridge, Mass., 2004.

\bibitem{sejdinovic2013equivalence}
Dino Sejdinovic, Bharath Sriperumbudur, Arthur Gretton, and Kenji Fukumizu.
\newblock Equivalence of distance-based and rkhs-based statistics in hypothesis
  testing.
\newblock {\em The Annals of Statistics}, pages 2263--2291, 2013.

\bibitem{Smyth1997}
Padhraic Smyth.
\newblock Clustering sequences with hidden markov models.
\newblock In {\em Advances in neural information processing systems}, pages
  648--654, 1997.

\bibitem{Strehl2002}
Alexander Strehl and Joydeep Ghosh.
\newblock Cluster ensembles --- a knowledge reuse framework for combining
  multiple partitions.
\newblock {\em J. Mach. Learn. Res.}, 3:583--617, 2002.

\bibitem{Stuetzle2003}
Werner Stuetzle.
\newblock Estimating the cluster tree of a density by analyzing the minimal
  spanning tree of a sample.
\newblock {\em Journal of Classification}, 20(1):25--47, 2003.

\bibitem{Stuetzle2010}
Werner Stuetzle and Rebecca Nugent.
\newblock A generalized single linkage method for estimating the cluster tree
  of a density.
\newblock {\em Journal of Computational and Graphical Statistics},
  19(2):397--418, 2010.

\bibitem{Thygesen2004}
Helene~H Thygesen and Aeilko~H Zwinderman.
\newblock Comparing transformation methods for dna microarray data.
\newblock {\em BMC bioinformatics}, 5(1):77, 2004.

\bibitem{OpenML2013}
Joaquin Vanschoren, Jan~N. van Rijn, Bernd Bischl, and Luis Torgo.
\newblock Openml: Networked science in machine learning.
\newblock {\em SIGKDD Explorations}, 15(2):49--60, 2013.

\bibitem{Vapnik19952}
Vladimir~N. Vapnik.
\newblock {\em The Nature of Statistical Learning Theory}.
\newblock Springer-Verlag, Berlin, Heidelberg, 1995.

\bibitem{Vega2011}
Sandro Vega-Pons and José Ruiz-Shulcloper.
\newblock A survey of clustering ensemble algorithms.
\newblock {\em International Journal of Pattern Recognition and Artificial
  Intelligence}, 25(03):337--372, 2011.

\bibitem{Vert2004}
Jean-Philippe Vert, Koji Tsuda, and Bernhard Sch{\"o}lkopf.
\newblock A primer on kernel methods.
\newblock {\em Kernel methods in computational biology}, 47:35--70, 2004.

\bibitem{wang2001efficient}
Fugao Wang and David~P Landau.
\newblock Efficient, multiple-range random walk algorithm to calculate the
  density of states.
\newblock {\em Physical review letters}, 86(10):2050, 2001.

\bibitem{Wang1997}
Wei Wang, Jiong Yang, Richard Muntz, et~al.
\newblock Sting: A statistical information grid approach to spatial data
  mining.
\newblock In {\em VLDB}, volume~97, pages 186--195, 1997.

\bibitem{Xuan2013}
Li~Xuan, Chen Zhigang, and Yang Fan.
\newblock Exploring of clustering algorithm on class-imbalanced data.
\newblock In {\em 8th International Conference on Computer Science and
  Education, ICCSE 2013}, pages 89--93, 04 2013.

\bibitem{Yeung-et-al-2001}
K.~Y. Yeung, C.~Fraley, A.~Murua, A.~E. Raftery, and W.~L. Ruzzo.
\newblock Model-based clustering and data transformations for gene expression
  data.
\newblock {\em Bioinformatics}, 17:977--987, 2001.

\end{thebibliography}

\end{document}